\documentclass[11pt]{article}

\newif\ifoldpdftex
\ifx\pdftexversion\undefined\else
  \ifnum\pdftexversion<140 \oldpdftextrue \fi
\fi
\ifoldpdftex
  \usepackage[nohyperref]{acl}
\else
  \usepackage[]{acl}
\fi

\usepackage{times}
\usepackage{latexsym}
\usepackage[T1]{fontenc}
\usepackage[utf8]{inputenc}
\IfFileExists{microtype.sty}{\usepackage{microtype}}{%
  \typeout{WARNING: microtype.sty not found; continuing without microtype.}}
\IfFileExists{inconsolata.sty}{\usepackage{inconsolata}}{%
  \typeout{WARNING: inconsolata.sty not found; continuing without inconsolata.}}
\usepackage{graphicx}

\usepackage{url}
\usepackage{enumitem}
\usepackage{xcolor}
\usepackage{booktabs}
\usepackage{amsmath}
\usepackage{float}
\IfFileExists{dblfloatfix.sty}{%
  \usepackage{dblfloatfix}%
}{%
  \typeout{WARNING: dblfloatfix.sty not found; continuing without dblfloatfix.}%
}
\renewcommand{\bottomfraction}{0.5}
\usepackage{listings}
\lstdefinestyle{prompt}{
  basicstyle=\ttfamily\fontsize{6.4pt}{7.2pt}\selectfont,
  breaklines=true,
  breakatwhitespace=true,
  breakindent=0pt,
  postbreak=\mbox{\textcolor{gray}{$\hookrightarrow$}},
  frame=none,
  xleftmargin=0pt,
  xrightmargin=0pt,
  aboveskip=0pt,
  belowskip=0pt,
  columns=fullflexible,
  keepspaces=true,
  showstringspaces=false,
}

\usepackage[most]{tcolorbox}
\usepackage{needspace}
\definecolor{promptbarbg}{HTML}{3F3F3F}
\newtcolorbox{promptbox}[1]{
  enhanced,
  breakable,  %
  colback=white, colframe=promptbarbg,
  colbacktitle=promptbarbg, coltitle=white,
  fonttitle=\bfseries\small, halign title=center,
  boxrule=0.6pt, arc=3pt,
  left=5pt, right=5pt, top=4pt, bottom=4pt,
  title={#1},
}

\definecolor{yang}{HTML}{2da02c}

\title{Disentangling Similarity and Relatedness in Topic Models}

\author{
  \parbox{\textwidth}{\centering
    \textbf{Hanlin Xiao}\textsuperscript{1,2},
    \textbf{Yang Wang}\textsuperscript{2},
    \textbf{Mauricio A \'Alvarez}\textsuperscript{2},
    \textbf{Rainer Breitling}\textsuperscript{1,3}\thanks{Corresponding author.}
  \\ \vspace{0.4cm}
    \textnormal{\textsuperscript{1}Manchester Institute of Biotechnology, The University of Manchester, UK} \\
    \textnormal{\textsuperscript{2}Department of Computer Science, The University of Manchester, UK} \\
    \textnormal{\textsuperscript{3}Department of Chemistry, The University of Manchester, UK}
  }
}

\newcommand{\atlasAppendixLegend}{%
1~=~\texttt{bertopic}, 2~=~\texttt{lsi}, 3~=~\texttt{lda}, 4~=~\texttt{fastopic}, 5~=~\texttt{etm-fresh}, 6~=~\texttt{ctm}, 7~=~\texttt{CT2V}, 8~=~\texttt{etm-bert}, 9~=~\texttt{prodlda}, 10~=~\texttt{nmf}, 11~=~\texttt{neurallda}, 12~=~\texttt{hdp}, 13~=~\texttt{cwtm}, 14~=~\texttt{ecrtm}, 15~=~\texttt{random}, 16~=~\texttt{T\&M}, 17~=~\texttt{bertkt}, 18~=~\texttt{tntm}%
}

\begin{document}
\maketitle

\begin{abstract}
The recent success of large pre-trained language models (PLMs) has motivated their integration into topic modeling. However, PLM-augmented topic models differ from classical co-occurrence models such as Latent Dirichlet Allocation (LDA) not only in performance, but also in the kind of semantic structure they capture. We formalize this distinction along two psycholinguistic axes: \textit{thematic relatedness} ({dog}/{bone}) and \textit{taxonomic similarity} ({dog}/{wolf}). To measure both axes over topic words, we build a large synthetic benchmark of word pairs via LLM annotation and train a neural scorer on it. Across multiple corpora and model families, the scorer places model families at distinct positions in the joint similarity--relatedness space. The two scores are able to predict downstream task performance: tasks requiring similarity benefit from similarity-rich topics, whereas tasks requiring relatedness benefit from the converse, and excess on either axis degrades performance on tasks aligned with the opposing axis. Neither axis is uniformly beneficial; measuring both yields a practical, model-agnostic diagnostic for topic semantic structure.\footnote{Code and data will be made publicly available upon acceptance.}  

\end{abstract}

\section{Introduction}

    Given a corpus of documents, topic models discover underlying thematic structure: \textit{topics}, each represented as a distribution over words that cohere around a common theme. For instance, \{{baseball}, {field}, {audience}, {win}\} suggests a topic of baseball games. Because of their scalability and interpretability, topic models have long been a standard tool for exploring large text collections~\citep{blei2012probabilistic,Boyd-Graber:Hu:Mimno-2017,grimmer2013text}, with applications in computational social science~\citep{egger2022topic,hansen2018transparency}, political science~\citep{barbera2019who,suter2025when}, digital humanities~\citep{tangherlini2013trawling}, and scientific literature mining~\citep{liu2021tracing}. Even in the current era of large language models (LLMs), they remain an attractive tool for corpus-level analysis: LLM-based topic extraction requires costly per-document inference~\citep{mu2024large,pham2024topicgpt}, whereas topic models process millions of documents at a small fraction of the time and computational cost while exposing corpus-wide statistics directly. Traditional probabilistic topic models, such as Latent Dirichlet Allocation (LDA)~\citep{blei2003lda}, rely on word co-occurrence statistics to represent each topic as a multinomial distribution over words and each document as a mixture of topics. More recently, the advent of contextual encoders and representation learning methods~\citep{devlin2019bert, liu2019roberta, pennington2014glove, openai2024embedding} has led to a new paradigm of PLM-augmented topic models~\citep{wu2024survey,grootendorst2022bertopic,wu2023ecrtm,wu2024fastopic,reuter2024tntm}. For example, ETM~\citep{dieng2020etm} employs a variational autoencoder architecture in which each topic is a vector in the word embedding space and per-topic word distributions are given by the softmax of topic--word embedding inner products, while other approaches, such as BERTopic~\citep{grootendorst2022bertopic}, cluster sentence-transformer embeddings and apply frequency weighting to extract topic candidates. PLM-augmented topic models have demonstrated strong performance on standard evaluation metrics such as topic diversity, coherence~\citep{wu2024survey}, and embedding-based measures~\citep{aletras2013distcoh,terragni2021webtsm}.

    However, existing evaluation metrics overlook a fundamental difference between these two families of models: probabilistic topic models tend to capture \emph{thematic relatedness}, a complementary association of functionally related words (e.g., \emph{coffee}--\emph{mug}: a mug is used to drink coffee, so the two are thematically related even though they refer to completely different objects), as they rely on co-occurrence statistics; whereas PLM-augmented topic models, by using embedding proximity, additionally capture \emph{taxonomic similarity}, substitutional closeness between categorically related words (e.g., \emph{coffee}--\emph{beer}: both are beverages and thus taxonomically similar, yet they play no complementary role in a shared scenario and are not thematically related). This distinction is well established in the psycholinguistic and NLP communities~\citep{agirre2009study,hill2015simlex,mirman2012individual,landrigan2016taxonomic}, and can be readily observed by inspecting topic examples from models such as BERT-KT~\citep{sia2020tired} (Appendix~\ref{app:bertkt}, Table~\ref{tab:lda_bertkt_example_topics}); however, no existing measure directly quantifies this two-dimensional distinction. Without it, this fundamental distinction goes unrecognized in model evaluation, further obscuring how model choice translates to downstream task performance. 

    In this paper, we address this gap by training a two-dimensional neural scorer for word pair similarity and relatedness, using a large-scale LLM-annotated synthetic dataset and static word embeddings. Using this scorer, we conduct a comprehensive survey of topic models across multiple corpora, revealing a consistent pattern among different model families. This distinction is predictive: the gap between the two scores consistently predicts downstream performance across task families, and excess on the opposing axis actually degrades performance when the semantic requirement is misaligned.
    
    The main contributions of this paper include:
\begin{itemize}
    \item We introduce a novel two-dimensional evaluation framework that explicitly disentangles taxonomic similarity from thematic relatedness in topic models, addressing a fundamental gap in how the semantic structure of topics is evaluated.
    
    \item We curate a 71K-pair similarity--relatedness dataset (${\sim}100{\times}$ larger than prior dual-annotated benchmarks) via sampling from lexical resources and a carefully prompted three-LLM ensemble, and train a neural scorer that quantifies both dimensions consistently. 
    
    \item We conduct a comprehensive evaluation of topic model families across multiple corpora, revealing a stable separation along the two axes, and showing that the axes act antagonistically on downstream task performance.
\end{itemize}

\section{Related Work}
\subsection{Topic Models and Evaluation}
Topic modeling has evolved through several generations: \textit{probabilistic generative} models such as LDA~\citep{blei2003lda} and its extensions (e.g., HDP~\citep{teh2006hdp}, CTM~\citep{lafferty2005correlated}); \textit{neural variational} approaches~\citep{srivastava2017avitm, miao2016nvdm}, which replace Dirichlet sampling with VAE-based inference; and \textit{PLM-augmented} topic models~\citep{grootendorst2022bertopic, bianchi2021ctm, reuter2024tntm}, which anchor the topics to pretrained language model embedding spaces such as BERT~\citep{devlin2019bert} and GloVe~\citep{pennington2014glove}. Standard evaluation relies on coherence metrics (UCI, UMass, NPMI)~\citep{mimno2011coherence,roder2015topiccoherence} that assess word co-occurrence patterns, diversity metrics (e.g., topic diversity and topic uniqueness)~\citep{terragni2021octis,wu2024topmost} and embedding-based methods that measure the quality in the embedding space~\citep{aletras2013distcoh,rahimi2024contextualized}. Prior work has also noted that coherence scores do not reliably correlate with human judgment~\citep{chang2009reading,lau2014coherence}. While topic models have been evaluated on downstream tasks such as document classification and information retrieval~\citep{mcauliffe2007supervised,Boyd-Graber:Hu:Mimno-2017}, these evaluations have not connected performance differences to the semantic structure of the topics themselves. 

\subsection{Similarity vs.\ Relatedness}

In this study, we adopt a definition of taxonomic similarity and thematic relatedness grounded in the psycholinguistic and NLP literature~\citep{mirman2012individual,landrigan2016taxonomic,estes2011thematic}. Taxonomic similarity reflects shared categorical features or defining properties, capturing a substitutional sense in which two words can replace one another in many contexts, as with synonyms or co-hyponyms. Thematic relatedness, by contrast, reflects a complementary functional association: words in the same event or scene that play different, complementary roles, as with \emph{coffee}--\emph{mug} or \emph{doctor}--\emph{hospital}.

This thematic notion of relatedness is different from the broader notion of distributional relatedness employed by many existing benchmarks~\citep{finkelstein2002placing,bruni2014multimodal}, under which any pair of words that share contexts is considered related. The two notions diverge on antonyms: pairs such as \emph{hot}--\emph{cold} attain high distributional relatedness because they occur in similar contexts~\citep{mohammad2013lexicalcontrast,ono2015antonym}, yet they are moderately to weakly thematically related, since they do not play complementary roles within a shared scene. Existing benchmarks largely target a single dimension: similarity-oriented resources such as SimLex-999~\citep{hill2015simlex} and RG-65~\citep{rubenstein1965contextual} isolate taxonomic similarity, whereas WordSim-353~\citep{finkelstein2002placing} and MEN~\citep{bruni2014multimodal} measure a broader distributional relatedness that conflates the two senses; TxThmNorms~\citep{landrigan2016taxonomic} and \citet{agirre2009study} provide signal on both dimensions but remain small and noisy. This motivates the two-dimensional scorer developed in this paper.

There are methods predicting each dimension separately using distributional methods~\citep{turney2010frequency,pennington2014glove,bojanowski2017enriching}, contextual PLM embeddings (e.g., BERT-based representations)~\citep{devlin2019bert,gupta2021obtaining,gao2021simcse}, and knowledge-based measures (e.g., WordNet and Wikipedia-based relatedness)~\citep{budanitsky2006evaluating,resnik1995using,gabrilovich2007computing}, but none jointly predict both on a continuous two-dimensional scale.

\section{Method}

\subsection{Similarity--Relatedness Scorer}

\subsubsection{Training Data Construction}
Building a robust similarity--relatedness neural scorer requires high-quality training data that distinguishes these two overlapping yet distinct semantic dimensions. Various word-pair datasets have been curated to evaluate similarity and relatedness~\citep{rubenstein1965contextual,finkelstein2002placing,agirre2009study,bruni2014multimodal,hill2015simlex}, but they typically focus on a single dimension and suffer from systematic conflation of the two (e.g., Table~\ref{tab:word_pairs}, Appendix~\ref{app:inconsistent}). Moreover, combining all existing manually annotated datasets yields only approximately 9,000 word pairs of noisy data, which is insufficient to fully exploit the capacity of a neural scoring model.

To address these limitations, we sampled from existing lexical databases and employed an LLM Imputation~\citep{schick2021generating,west2022symbolic} approach to curate large-scale training data. By targeting specific word relations, including shared synsets and co-hyponyms in WordNet~\citep{miller1995wordnet}, associative edges (\textit{RelatedTo}, \textit{UsedFor}, \textit{PartOf}, \textit{AtLocation}, \textit{Causes}) in ConceptNet~\citep{speer2017conceptnet}, cosine-nearest neighbors in the BERT embedding space~\citep{devlin2019bert}, and controlled negatives from cross-domain and cross-POS pairings, we obtained candidate word pairs whose relationships span the full spectrum of similarity and relatedness. Candidate pairs were annotated by averaging scores from an ensemble of three large language models (DeepSeek-V4~\citep{deepseekai2026v4}, GPT-4.1-mini~\citep{openai2025gpt41}, and Qwen-Turbo~\citep{qwen2025qwenturbo}). Each annotator was prompted with the definitions and explicit scoring anchors. Three design choices behind this pipeline are validated and ablated by evaluation on the external, human-rated TxThmNorms benchmark~\citep{landrigan2016taxonomic} in Appendix~\ref{app:annotator_ablation}, including the prompt format, the choice of annotators, and the use of an ensemble.

Through this pipeline, we curated a dataset of 71,346 word pairs that offers balanced scoring across the similarity--relatedness spectrum and broad lexical diversity through its multi-source design. Still, the dataset could be expanded to potentially further improve the scorer (Appendix~\ref{app:scaling_law}).

\subsubsection{Neural Network Architecture and Training}
We train a neural scoring function that takes a word pair as input and jointly predicts similarity and relatedness scores, optimized with a sum of per-dimension masked MSE losses $\mathcal{L} = \mathcal{L}_{\text{sim}} + \mathcal{L}_{\text{rel}}$. The two heads share a common trunk but are decoupled in the loss, so each dimension is learned purely from its own labels.

Each word is embedded with OpenAI's sentence encoder text-embedding-3-large~\citep{openai2024embedding} applied to a short prompt template, following the prompted-encoding scheme of \citet{ranjan2026oneword}. From a pair of word embeddings $(\mathbf{e}_1, \mathbf{e}_2)$ we construct a joint representation by concatenating four components, $[\,\mathbf{e}_1;\; \mathbf{e}_2;\; |\mathbf{e}_1 - \mathbf{e}_2|;\; \mathbf{e}_1 \odot \mathbf{e}_2\,]$~\citep{mou2016tbcnn,conneau2017infersent}, augmented with an 8-dimensional WordNet-derived feature vector that injects explicit taxonomic structure. The resulting representation is passed through a multilayer perceptron with two scalar output heads. We ablate the input embedding, the contribution of WordNet features, and the network architecture against a range of alternative encoders and architectural variants in Appendix~\ref{app:ablation}; full architectural and hyperparameter settings are in Appendix~\ref{app:neural_scorer}.

\begin{figure*}
    \centering
    \includegraphics[width=\linewidth]{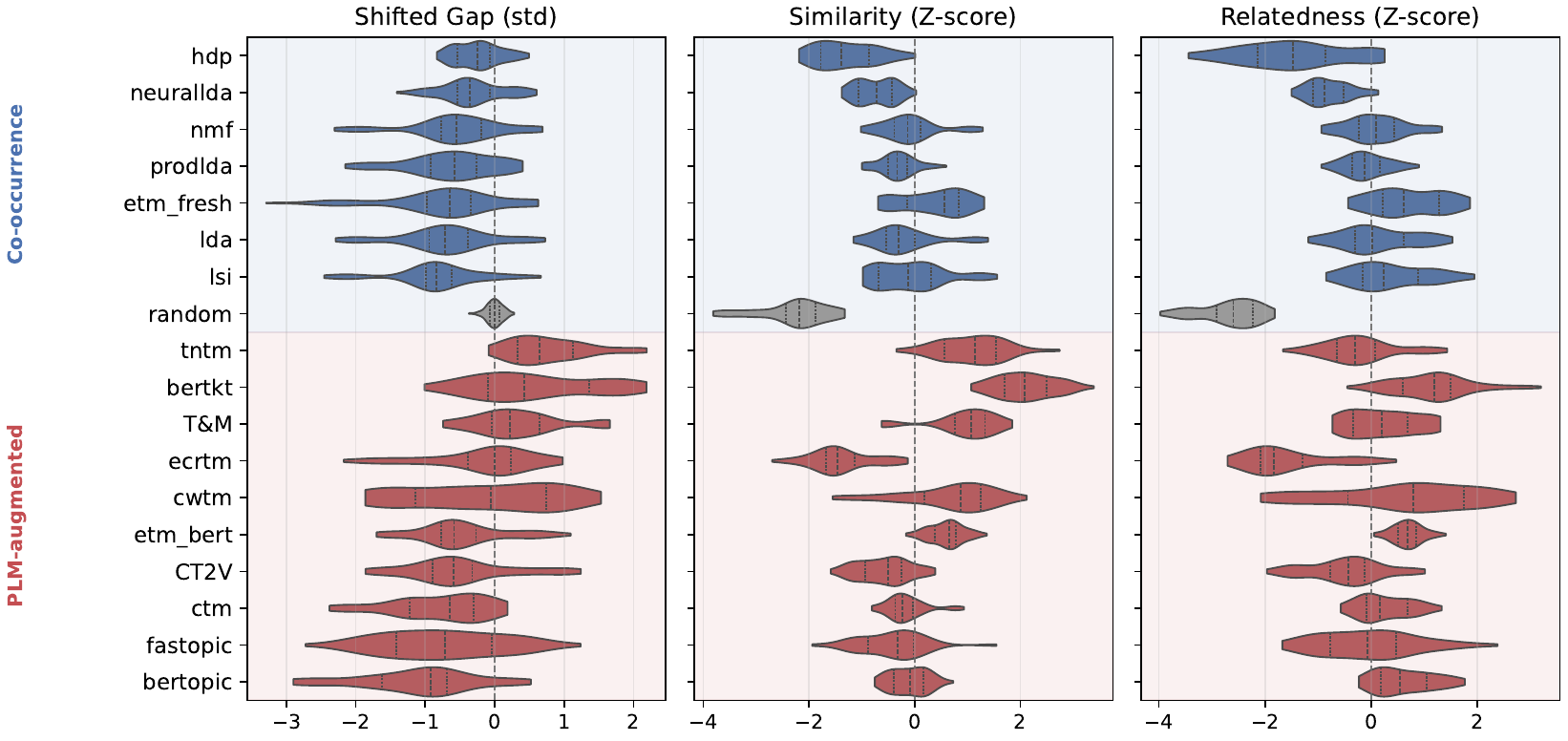}
    \caption{Topic model atlas pooled across all nine corpora. Each violin shows the distribution of per-topic scores over 9 corpora $\times$ 10 seeds $\times$ 20 topics for one of 17 models, colored by model family (Appendix~\ref{app:model_inventory} for the full model inventory). Left: shifted gap, defined as normalized similarity minus normalized relatedness, shifted so a random-word baseline sits at zero; positive values lean similarity, negative lean relatedness. Centre and right: per-corpus Z-scores on each axis. Per-corpus breakdowns in Appendix~\ref{app:atlas_viz}.}
    \label{fig:AtlasReuters}
\end{figure*}

\subsection{Topic Model Atlas}

\subsubsection{Topic Models and Corpora}

We apply the trained scorer to a comprehensive evaluation of 17 topic models and a random baseline on nine corpora. The models span several methodological families: co-occurrence-based probabilistic models, neural VAE-based extensions, and embedding-clustering approaches based on PLMs; they can be grouped into two classes, co-occurrence-based and PLM-augmented. The corpora span scientific writing, newswire, and informal forum discussion; document lengths from short titles and headlines to full-length articles; and collections compiled over several decades, combining six established topic-modeling benchmarks with three larger, more recent additions. Full inventories with citations, category counts, and preprocessing settings are provided in Appendices~\ref{app:model_inventory} (Table~\ref{tab:atlas-models}) and~\ref{app:corpora} (Table~\ref{tab:corpus_stats}).

\subsubsection{Atlas Construction and Consistency Analysis}
For each (model, corpus) pair, we run the topic model with $k = 20$ topics and retain the top 10 words per topic, repeating 10 times with different random seeds. This yields up to 90 independent topic sets per model across the atlas (nine corpora $\times$ ten runs).

We then score each topic with the trained scorer: for its top 10 words, we compute the similarity and relatedness of all $\binom{10}{2} = 45$ word pairs and take the unweighted mean, giving one 2D score per topic. Aggregating upward yields two summary levels: for each (model, corpus) cell, the mean over its 20 topics $\times$ 10 runs together with the within-cell variance across topics and runs; and for each model overall, the mean across the nine corpora together with the cross-corpus variance.

To summarize a model's position along the similarity--relatedness spectrum, we define the \textit{shifted normalized gap}. Within each corpus $c$, we pool the raw similarity and relatedness scores of all topics from all non-random models and runs, and $z$-score standardize each axis using their mean and standard deviation; normalization is thus corpus-internal and model-agnostic. For a topic $t$ in corpus $c$, let $\tilde{s}_{t,c}$ and $\tilde{r}_{t,c}$ denote its per-corpus $z$-standardized similarity and relatedness, and let $\bar{g}^{\text{rand}}_c$ be the mean of $\tilde{s}_{t,c} - \tilde{r}_{t,c}$ over random-word topics (sampled uniformly from the corpus's own vocabulary) in $c$. The shifted normalized gap is then
\begin{equation}
g_{t,c} = (\tilde{s}_{t,c} - \tilde{r}_{t,c}) - \bar{g}^{\text{rand}}_c,
\end{equation}
so that the random-word baseline sits at zero. Positive values indicate a similarity-leaning profile relative to random; negative values, a relatedness-leaning one. To test whether this profile is stable across corpora, we compute Kendall's coefficient of concordance $W$~\citep{kendall1948rank} over the per-corpus model rankings.

Not all (model, corpus) cells could be obtained in practice: a small number suffered from topic collapse or prohibitive running times and were excluded (full list in Table~\ref{tab:omitted_models}, Appendix~\ref{app:omitted}). Despite these exclusions, 1{,}567 individual runs were successfully completed across the atlas, with per-corpus counts ranging from 158 to 180.

\subsection{Downstream Tasks}
The proposed measure serves as a diagnostic for selecting topic models for downstream deployment. We test this by evaluating its predictiveness on three practical scenarios in which topic models are used: event monitoring, document retrieval, and synonym-aware retrieval (query expansion). The first two reward models that capture co-occurrence structure; the third rewards models that capture categorical interchangeability. We compare the metric against established measures to assess whether it provides stronger guidance for matching topic models to downstream tasks.

\subsubsection{Event Monitoring}

Event monitoring asks whether a topic representation can route incoming documents to the right thematic bucket, the core operation behind newswire triage, content moderation, and trend tracking~\citep{Boyd-Graber:Hu:Mimno-2017}. We operationalize this as a centroid-based retrieval task: for each labeled category in the training set, we average its documents' topic vectors into a category centroid; at test time, each document is embedded as a topic vector and ranked against all centroids. Performance is measured by Hit and Recall. Because the task rewards recognizing which words form coherent themes, it favors models that capture this co-occurrence structure, which our measure quantifies as relatedness. We evaluate on every label-bearing corpus in the atlas; full metric and corpus details, along with per-model results, are reported in Appendix~\ref{app:task_tables}.

\subsubsection{Category Retrieval}
Category retrieval asks whether documents on the same topic land near each other in topic space, the core operation behind recommendation and case-based retrieval. We operationalize this as nearest-neighbor consistency: each document is embedded as a topic vector, and for every document we retrieve its 10 nearest neighbors and check whether their category labels match the query. Performance is measured by the same-category rate (SCR), the fraction of the 10 nearest neighbors sharing at least one category label with the query. Both retrieval tasks reward relatedness and rely on the same topic-space geometry, but probe it at different scales (centroid versus local neighborhood); we treat them as complementary checks on the same hypothesis rather than independent tests.

\subsubsection{Synonym-Aware Retrieval}

The third task asks whether synonymous queries surface the same documents, the core operation behind query expansion and synonym-aware search. We curate a set of strict synonym pairs per corpus (e.g., \textit{increase}/\textit{rise}; full inventory in Appendix~\ref{app:synonym_pairs}) and, for each word, compute the conditional probability $P(w|d)$ of word $w$ given document $d$ by marginalizing over topics using the model's native topic--word distribution. We then take the top-10 highest-scoring documents per word and measure the overlap between the two ranked lists by Jaccard similarity. Because synonyms are semantically interchangeable, a model that captures categorical similarity should induce similar document rankings for the two members of a pair. The task requires no ground-truth labels, so we evaluate it on all nine corpora; it does require access to the native topic--word distribution, so we include only the model families that expose one (Appendix~\ref{app:taskc_per_corpus}; Figure~\ref{fig:taskC} shows the cross-corpus fit).

\subsubsection{Regression Analysis}
To holistically assess whether the similarity--relatedness metric predicts downstream performance, we regress per-(model, corpus) task performance on the model-level scores using pooled cross-corpus regressions with corpus fixed effects (a per-corpus intercept that absorbs baseline differences). The analysis is intended as an overview of predictive behavior across the atlas, not as a per-corpus diagnostic. We report two complementary specifications. The first enters each predictor on its own and summarizes it by partial $R^2$; we compare the similarity score, the relatedness score, and their gap against established metrics ($C_V$ coherence and topic diversity). The second is a joint model that enters similarity and relatedness together, so that each coefficient is estimated controlling for the other. Statistical significance is reported for every predictor. The full specification (panel construction, within-corpus $z$-scoring, partial-$R^2$ definition) is given in Appendix~\ref{app:regression_method}.

\section{Results}
This section reports the empirical findings in three parts. We first validate the similarity--relatedness scorer on held-out data and the external TxThmNorms benchmark, showing that it captures both dimensions consistently and keeps them disentangled. Next, we show that each model's similarity--relatedness profile is stable across corpora, reflecting an intrinsic property of the model. Finally, we use cross-corpus regressions to test whether the two scores predict downstream task performance, benchmarking against established measures, and find that the two dimensions act antagonistically across task families. Per-model task results are detailed in Appendix~\ref{app:task_tables}, full per-corpus atlas visualizations in Appendix~\ref{app:atlas_viz}, and per-corpus Synonym-Aware Retrieval performance in Appendix~\ref{app:taskc_per_corpus}.

\begin{table}[h]
\centering
\caption{Spearman rank correlation ($\rho_s$) with human similarity (Sim.) and relatedness (Rel.) ratings on the 659-pair TxThmNorms benchmark. Baselines: cosine similarity over single-word BERT and OpenAI {text-embedding-3-large} embeddings, and WordNet Wu--Palmer similarity. }
\label{tab:scorer_eval}
\begingroup
\small
\setlength{\tabcolsep}{3pt}
\renewcommand{\arraystretch}{0.95}
\begin{tabular}{@{}lcc@{}}
\toprule
Model & Sim.\ ($\rho_s$) & Rel.\ ($\rho_s$) \\
\midrule
Neural Scorer     & \textbf{0.667}  &  \textbf{0.506} \\
BERT~\citep{devlin2019bert}       & 0.229  &  0.239 \\
OpenAI~\citep{openai2024embedding}     & 0.347  &  0.182 \\
WordNet W--P~\citep{wu1994verb} & 0.620  & $-0.374$ \\
\bottomrule
\end{tabular}
\endgroup
\end{table}

\subsection{Scorer Validation}
\label{sec:results_scorer}

We first validate the scorer on a held-out split of the synthetic dataset. The loss curve on the held-out split decreases during training and stabilizes without degradation (Figure~\ref{fig:training}, Appendix~\ref{app:neural_scorer}). Externally, we evaluate on TxThmNorms~\citep{landrigan2016taxonomic}, a human-annotated benchmark covering both dimensions. Table~\ref{tab:scorer_eval} compares the scorer against other baselines. Our neural scorer substantially outperforms all baselines on both dimensions, avoiding pure embeddings' conflation of the two and WordNet's failure to capture thematic relatedness.

\subsection{Topic Model Atlas}

Figure~\ref{fig:AtlasReuters} presents the atlas of all 17 models pooled across the nine corpora along the similarity, relatedness, and shifted-gap axes. No single axis cleanly separates the two classes (co-occurrence-based, PLM-augmented). Each axis captures one facet of topic semantics, and class membership is only loosely predictive of any single facet. Reading the three panels jointly exposes class-conditional tendencies.

\begin{table}[h]
\centering
\caption{Cross-corpus rank agreement of topic models across nine datasets, quantified by Kendall's coefficient of concordance $W$; all three axes attain $p < 0.001$. }
\label{tab:kendall_w}
\small
\begin{tabular}{lc}
\toprule
\textbf{Evaluation Axis} & \textbf{Kendall's $W$} \\
\midrule
Similarity & 0.768 \\
Relatedness & 0.743 \\
Shifted Normalized Gap & 0.732 \\
\bottomrule
\end{tabular}
\end{table}

\paragraph{Co-occurrence-based models} lean toward relatedness, consistent with their reliance on corpus co-occurrence statistics, and accordingly occupy the relatedness-anchored end of the shifted-gap spectrum. Two members, \texttt{hdp} and \texttt{neurallda}, depart from this pattern by scoring low on \emph{both} axes; the shifted gap then places them adjacent to the PLM-augmented block, not because their topics are similarity-anchored but because relatedness has collapsed. This degenerate regime is undetectable from the gap alone and motivates reporting all three axes jointly.

\paragraph{PLM-augmented models} exhibit large within-class variance. Several remain heavily shaped by corpus co-occurrence (BERTopic through c-TF-IDF reranking, CombinedTM through a bag-of-words decoder) and consequently sit closer to the co-occurrence-based models than to the rest of their own class, while others, such as BERT-KT and T\&M, focus on the natural clustering of embeddings, which tends more toward similarity. A PLM-augmented model's position on the spectrum is thus determined by which component, embeddings or co-occurrence-based techniques, dominates topic construction, rather than by the mere presence of pretrained embeddings.

    \begin{figure}[htbp]
    \centering
    \includegraphics[width=\linewidth]{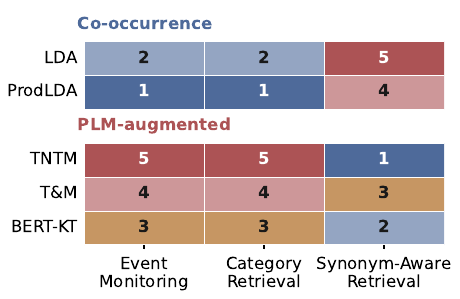}
    \caption{Per-task rank (1~$=$~best, 5~$=$~worst) of five models on three downstream tasks, grouped by class; cell color encodes rank (blue $=$ best, red $=$ worst). The double dissociation is visible: co-occurrence-based models lead on Event Monitoring and Category Retrieval but trail on Synonym-Aware Retrieval; PLM-augmented models reverse the pattern.}
    \label{fig:dtask}
    \end{figure}

\paragraph{Cross-corpus stability.} Although the two classes overlap on the spectrum, the model rankings they induce are stable across corpora. Kendall's coefficient of concordance $W$~\citep{kendall1948rank} attains $W = 0.73$--$0.77$ on all three axes (Table~\ref{tab:kendall_w}). Given the substantial heterogeneity in domain, vocabulary, and document length across the nine corpora, this level of agreement indicates that a model's position along the similarity--relatedness axes reflects an intrinsic property of the model rather than a corpus-specific artefact. Per-corpus breakdowns are provided in Appendix~\ref{app:atlas_viz}.

\subsection{Downstream Task Prediction}

Figure~\ref{fig:dtask} visualizes per-model performance across the three tasks for a representative subset of models, with full per-model results reported in Appendix~\ref{app:task_tables}. The split predicted by the atlas is immediately visible: co-occurrence-based models dominate the relatedness-favoring Event Monitoring and Category Retrieval, whereas PLM-augmented models lead on the similarity-favoring Synonym-Aware Retrieval.

Table~\ref{tab:single_factor} quantifies this pattern through a cross-corpus pooled fixed-effects regression, comparing the proposed two-dimensional metric (similarity, relatedness, and their gap) against existing intrinsic measures. In the single-factor panel, the gap is the only predictor that is strong and significant on all three tasks; coherence and diversity each cover one task family only, and raw similarity and relatedness each predict the matched task but collapse on the opposing one. To understand why the gap succeeds where its constituents individually fail, we turn to the two-factor panel. Entered jointly, the two coefficients take \emph{opposite} signs on every task. The practical implication is that a model's lean, not its absolute level on either axis, predicts task-specific performance: for a given downstream task, a model whose profile is purified toward the task-aligned axis outperforms one that scores high on both.

\section{Discussion}

\paragraph{Two Axes Act Antagonistically}
The opposite-sign coefficients follow from what each task requires. Event identification depends on words that play distinct, complementary roles in a shared scene; high taxonomic similarity may introduce categorically close words drawn from entirely different scenarios, blurring the event. Synonym-aware retrieval depends on the opposite property, words being interchangeable, which a relatedness-leaning topic does not provide. The two axes therefore pull in opposite directions, and a model's lean between them determines which task it serves.

\paragraph{Applications}
The two-dimensional measure exposes structural trade-offs invisible to coherence and diversity, recasting model selection as an explicit choice over the kind of semantic structure captured. In embedding-augmented VAE topic models, the profile tracks how the embedding component's weight tilts topics toward taxonomic similarity, guiding architectural and hyperparameter decisions. Its link to downstream performance enables task-aware matching: retrieval-augmented generation~\citep{lewis2020retrieval, tran2024enhancing} may favor relatedness-inclined models for recall and similarity-inclined ones for reranking. The scorer itself also generalizes to arbitrary word sets, e.g., keyword extraction evaluation~\citep{wu2024kpeval} or assessing cluster cohesion.

\paragraph{Relation to Existing Metrics}
Coherence and embedding-based measures are behavioral: they quantify how topic words pattern in corpus co-occurrence or embedding proximity. Our scorer targets a different level of description, the kind of semantic relation that organizes the words, drawn from external lexical knowledge and resolved into two orthogonal axes. The two families overlap in part: coherence parallels our relatedness axis, since both reflect thematic association, but they remain distinct in what they measure. The two types of metrics are therefore complementary, and reporting them jointly provides a holistic view of topic structure that supports the downstream predictions.

\begin{table}[h]
\centering
\caption{Cross-corpus pooled regression of downstream performance. \textbf{Top:} single-factor partial $R^2$. \textbf{Bottom:} joint two-factor coefficients ($z$-scored within corpus).}
\label{tab:single_factor}
\small
\setlength{\tabcolsep}{3pt}
\renewcommand{\arraystretch}{1.18}
\begin{tabular}{@{}lccc@{}}
\toprule
 & \multicolumn{2}{c}{\emph{Relatedness}} & \emph{Similarity} \\
 \cmidrule(lr){2-3} \cmidrule(lr){4-4}
\textbf{Predictor} & Event Mon. & Cat. Retr. & Syn-Aware Retr. \\
\midrule
\multicolumn{4}{@{}l}{\emph{Single-factor partial $R^2$}} \\
Similarity        & .00          & .06          & \textbf{.53} \\
Relatedness       & \textbf{.24} & \textbf{.42} & .03          \\
\textbf{Gap}      & \textbf{.64} & \textbf{.64} & \textbf{.77} \\
$C_V$ Coherence   & \textbf{.49} & \textbf{.41} & .08          \\
Diversity         & .03          & .11          & \textbf{.44} \\
\midrule
\multicolumn{4}{@{}l}{\emph{Two-factor joint coefficients}} \\
$\beta_{\text{sim}}$ & $\mathbf{-1.01}$ & $\mathbf{-0.88}$ & $\mathbf{+0.34}$ \\
$\beta_{\text{rel}}$ & $\mathbf{+1.31}$ & $\mathbf{+1.37}$ & $\mathbf{-0.19}$ \\
\bottomrule
\end{tabular}\\[3pt]
{\scriptsize\raggedright Retrieval tasks: $n{=}46$ ($n{=}34$ for coherence, undefined on phrase-topic models). Synonym-Aware Retrieval: $n{=}38$ ($n{=}33$ for coherence). \textbf{Bold}: $p<.001$.\par}
\end{table}

\section{Conclusion}
This work introduces a two-dimensional evaluation framework that disentangles \textit{taxonomic similarity} from \textit{thematic relatedness} in topic model outputs. We train a neural scorer on a large-scale, LLM-annotated synthetic dataset and apply it to an atlas of 17 topic models across 9 corpora, finding that the similarity--relatedness profile is a stable, intrinsic property of each model class. The distinction is also predictive: in pooled cross-corpus regressions, the gap between the two scores is the only intrinsic metric that predicts performance across both relatedness-favoring and similarity-favoring task families, and the two axes act antagonistically once their shared topic-quality component is removed. Beyond a new metric, our results reframe topic model evaluation as a question of \textit{what kind} of semantic structure a model captures, enabling task-aware model selection and principled tuning of embedding-augmented architectures.

\section*{Limitations}

\textbf{Configuration choices.} The atlas fixes several evaluation settings: $k = 20$ topics per model, the top 10 words per topic, 10 random seeds per (model, corpus) cell, and a unified preprocessing pipeline shared across all corpora. Within a topic, the two scores are aggregated as an unweighted mean over all 45 word pairs, without weighting by topic--word probability. These settings follow established evaluation practice and keep the 1,567 runs comparable, but we did not sweep them; how the similarity--relatedness profile responds to topic granularity, word-list depth, or probability-weighted aggregation is left to future work.

\noindent\textbf{Component choices.} Beyond the fixed settings above, the pipeline contains several interchangeable components for which we adopted one reasonable instantiation each: the encoder underlying each PLM-augmented model, the scorer's input embedding and architecture (selected in Appendix~\ref{app:ablation} rather than exhaustively searched), and the comparator metrics in Table~\ref{tab:single_factor} ($C_V$ coherence and topic diversity, out of a wider inventory of coherence variants and embedding-based measures). We did not enumerate combinations of these components. Within the scope we do cover, however, the reported pattern is robust: the model rankings induced by both axes are concordant across nine heterogeneous corpora ($W = 0.73$--$0.77$, Table~\ref{tab:kendall_w}), indicating that the separation between the two axes and its downstream predictiveness are properties of the models rather than artefacts of a particular configuration. Substituting components would likely shift where individual models sit on the spectrum, rather than overturn the pattern itself.

\section*{Acknowledgements}
This work was supported by the Engineering and Physical Sciences Research Council [EP/Y030826/1].

\bibliography{publication/main/2_atlas/model_references,refs}

@article{blei2003lda,
  author  = {Blei, David M. and Ng, Andrew Y. and Jordan, Michael I.},
  title   = {Latent {D}irichlet Allocation},
  journal = {Journal of Machine Learning Research},
  volume  = {3},
  pages   = {993--1022},
  year    = {2003}
}

@article{deerwester1990lsi,
  author  = {Deerwester, Scott and Dumais, Susan T. and Furnas, George W. and
             Landauer, Thomas K. and Harshman, Richard},
  title   = {Indexing by Latent Semantic Analysis},
  journal = {Journal of the American Society for Information Science},
  volume  = {41},
  number  = {6},
  pages   = {391--407},
  year    = {1990}
}

@article{teh2006hdp,
  author  = {Teh, Yee Whye and Jordan, Michael I. and Beal, Matthew J. and Blei, David M.},
  title   = {Hierarchical {D}irichlet Processes},
  journal = {Journal of the American Statistical Association},
  volume  = {101},
  number  = {476},
  pages   = {1566--1581},
  year    = {2006}
}

@article{lee1999nmf,
  author  = {Lee, Daniel D. and Seung, H. Sebastian},
  title   = {Learning the Parts of Objects by Non-negative Matrix Factorization},
  journal = {Nature},
  volume  = {401},
  pages   = {788--791},
  year    = {1999}
}

@inproceedings{srivastava2017avitm,
  author    = {Srivastava, Akash and Sutton, Charles},
  title     = {Autoencoding Variational Inference for Topic Models},
  booktitle = {International Conference on Learning Representations (ICLR)},
  year      = {2017}
}

@article{dieng2020etm,
  author  = {Dieng, Adji B. and Ruiz, Francisco J. R. and Blei, David M.},
  title   = {Topic Modeling in Embedding Spaces},
  journal = {Transactions of the Association for Computational Linguistics},
  volume  = {8},
  pages   = {439--453},
  year    = {2020}
}

@inproceedings{bianchi2021ctm,
  author    = {Bianchi, Federico and Terragni, Silvia and Hovy, Dirk},
  title     = {Pre-training is a Hot Topic: Contextualized Document Embeddings
               Improve Topic Coherence},
  booktitle = {Proceedings of the 59th Annual Meeting of the Association for
               Computational Linguistics and the 11th International Joint
               Conference on Natural Language Processing (Volume 2: Short Papers)},
  pages     = {759--766},
  year      = {2021}
}

@article{grootendorst2022bertopic,
  author  = {Grootendorst, Maarten},
  title   = {{BERT}opic: Neural Topic Modeling with a Class-Based {TF-IDF} Procedure},
  journal = {arXiv preprint arXiv:2203.05794},
  year    = {2022}
}

@inproceedings{angelov2024top2vec,
  author    = {Angelov, Dimo and Inkpen, Diana},
  title     = {Topic Modeling: Contextual Token Embeddings Are All You Need},
  booktitle = {Findings of the Association for Computational Linguistics: EMNLP 2024},
  year      = {2024}
}

@inproceedings{wu2023ecrtm,
  author    = {Wu, Xiaobao and Dong, Xinshuai and Nguyen, Thong Thanh and Luu, Anh Tuan},
  title     = {Effective Neural Topic Modeling with Embedding Clustering Regularization},
  booktitle = {Proceedings of the 40th International Conference on Machine Learning (ICML)},
  year      = {2023}
}

@article{thompson2020clusters,
  author  = {Thompson, Laure and Mimno, David},
  title   = {Topic Modeling with Contextualized Word Representation Clusters},
  journal = {arXiv preprint arXiv:2010.12626},
  year    = {2020}
}

@inproceedings{fang2024cwtm,
  author    = {Fang, Zheng and He, Yulan and Procter, Rob},
  title     = {{CWTM}: Leveraging Contextualized Word Embeddings from {BERT}
               for Neural Topic Modeling},
  booktitle = {Proceedings of the 2024 Joint International Conference on
               Computational Linguistics, Language Resources and Evaluation
               (LREC-COLING)},
  pages     = {4273--4286},
  year      = {2024}
}

@article{reuter2024tntm,
  author  = {Reuter, Arik and Thielmann, Anton and Weisser, Christoph and
             S{\"a}fken, Benjamin and Kneib, Thomas},
  title   = {Probabilistic Topic Modeling with Transformer Representations},
  journal = {IEEE Transactions on Neural Networks and Learning Systems},
  year    = {2025},
  doi     = {10.1109/TNNLS.2025.3538262}
}

@inproceedings{wu2024fastopic,
  author    = {Wu, Xiaobao and Nguyen, Thong Thanh and Zhang, Delvin Ce and
               Wang, William Yang and Luu, Anh Tuan},
  title     = {{FAST}opic: Pretrained Transformer is a Fast, Adaptive, Stable,
               and Transferable Topic Model},
  booktitle = {Advances in Neural Information Processing Systems (NeurIPS)},
  year      = {2024}
}

@inproceedings{sia2020tired,
  title     = {Tired of topic models? {C}lusters of pretrained word embeddings make for fast and good topics too!},
  author    = {Sia, Suzanna and Dalmia, Ayush and Mielke, Sabrina J.},
  booktitle = {Proceedings of the 2020 Conference on Empirical Methods in Natural Language Processing ({EMNLP})},
  pages     = {1728--1736},
  year      = {2020},
  address   = {Online},
  publisher = {Association for Computational Linguistics},
  url       = {https://aclanthology.org/2020.emnlp-main.135/},
  doi       = {10.18653/v1/2020.emnlp-main.135}
}

@inproceedings{aletras2013distcoh,
  title     = {Evaluating topic coherence using distributional semantics},
  author    = {Aletras, Nikolaos and Stevenson, Mark},
  booktitle = {Proceedings of the 10th International Conference on Computational Semantics ({IWCS} 2013) -- Long Papers},
  pages     = {13--22},
  year      = {2013},
  address   = {Potsdam, Germany},
  publisher = {Association for Computational Linguistics},
  url       = {https://aclanthology.org/W13-0102/}
}

@inproceedings{terragni2021webtsm,
  title     = {Word embedding-based topic similarity measures},
  author    = {Terragni, Silvia and Fersini, Elisabetta and Messina, Enza},
  booktitle = {Natural Language Processing and Information Systems},
  editor    = {M{\'e}tais, Elisabeth and Meziane, Farid and Horacek, Helmut and Kapetanios, Epaminondas},
  series    = {Lecture Notes in Computer Science},
  volume    = {12801},
  pages     = {33--45},
  year      = {2021},
  publisher = {Springer, Cham},
  doi       = {10.1007/978-3-030-80599-9_4},
  url       = {https://doi.org/10.1007/978-3-030-80599-9_4}
}

@inproceedings{agirre2009study,
  title     = {A study on similarity and relatedness using distributional and {WordNet}-based approaches},
  author    = {Agirre, Eneko and Alfonseca, Enrique and Hall, Keith and Kravalova, Jana and Pa{\c{s}}ca, Marius and Soroa, Aitor},
  booktitle = {Proceedings of Human Language Technologies: The 2009 Annual Conference of the North American Chapter of the Association for Computational Linguistics},
  pages     = {19--27},
  year      = {2009},
  address   = {Boulder, Colorado},
  publisher = {Association for Computational Linguistics},
  url       = {https://aclanthology.org/N09-1003/}
}

@article{hill2015simlex,
  title     = {{SimLex-999}: evaluating semantic models with (genuine) similarity estimation},
  author    = {Hill, Felix and Reichart, Roi and Korhonen, Anna},
  journal   = {Computational Linguistics},
  volume    = {41},
  number    = {4},
  pages     = {665--695},
  year      = {2015},
  publisher = {MIT Press},
  doi       = {10.1162/COLI_a_00237},
  url       = {https://aclanthology.org/J15-4004/}
}

@article{mirman2012individual,
  title   = {Individual differences in the strength of taxonomic versus thematic relations},
  author  = {Mirman, Daniel and Graziano, Kristen M.},
  journal = {Journal of Experimental Psychology: General},
  volume  = {141},
  number  = {4},
  pages   = {601--609},
  year    = {2012},
  doi     = {10.1037/a0026451}
}

@article{landrigan2016taxonomic,
  title     = {Taxonomic and thematic relatedness ratings for 659 word pairs},
  author    = {Landrigan, Jon-Frederick and Mirman, Daniel},
  journal   = {Journal of Open Psychology Data},
  volume    = {4},
  number    = {1},
  pages     = {e2},
  year      = {2016},
  publisher = {Ubiquity Press},
  doi       = {10.5334/jopd.24},
  url       = {https://openpsychologydata.metajnl.com/articles/10.5334/jopd.24}
}

@inproceedings{miao2016nvdm,
  title     = {Neural variational inference for text processing},
  author    = {Miao, Yishu and Yu, Lei and Blunsom, Phil},
  booktitle = {Proceedings of the 33rd International Conference on Machine Learning ({ICML})},
  editor    = {Balcan, Maria Florina and Weinberger, Kilian Q.},
  series    = {Proceedings of Machine Learning Research},
  volume    = {48},
  pages     = {1727--1736},
  year      = {2016},
  publisher = {PMLR},
  url       = {https://proceedings.mlr.press/v48/miao16.html}
}

@inproceedings{mimno2011coherence,
  title     = {Optimizing semantic coherence in topic models},
  author    = {Mimno, David and Wallach, Hanna M. and Talley, Edmund and Leenders, Miriam and McCallum, Andrew},
  booktitle = {Proceedings of the 2011 Conference on Empirical Methods in Natural Language Processing ({EMNLP})},
  pages     = {262--272},
  year      = {2011},
  address   = {Edinburgh, Scotland, UK},
  publisher = {Association for Computational Linguistics},
  url       = {https://aclanthology.org/D11-1024/}
}

@inproceedings{roder2015topiccoherence,
  title     = {Exploring the space of topic coherence measures},
  author    = {R{\"o}der, Michael and Both, Andreas and Hinneburg, Alexander},
  booktitle = {Proceedings of the Eighth ACM International Conference on Web Search and Data Mining ({WSDM})},
  pages     = {399--408},
  year      = {2015},
  publisher = {Association for Computing Machinery},
  doi       = {10.1145/2684822.2685324},
  url       = {https://dl.acm.org/doi/10.1145/2684822.2685324}
}

@inproceedings{chang2009reading,
  title     = {Reading tea leaves: how humans interpret topic models},
  author    = {Chang, Jonathan and Gerrish, Sean and Wang, Chong and Boyd-Graber, Jordan and Blei, David M.},
  booktitle = {Advances in Neural Information Processing Systems},
  volume    = {22},
  pages     = {288--296},
  year      = {2009},
  publisher = {Curran Associates, Inc.},
  url       = {https://papers.nips.cc/paper_files/paper/2009/hash/f92586a25bb3145facd64ab20fd554ff-Abstract.html}
}

@inproceedings{lau2014coherence,
  title     = {Machine reading tea leaves: automatically evaluating topic coherence and topic model quality},
  author    = {Lau, Jey Han and Newman, David and Baldwin, Timothy},
  booktitle = {Proceedings of the 14th Conference of the European Chapter of the Association for Computational Linguistics ({EACL})},
  pages     = {530--539},
  year      = {2014},
  address   = {Gothenburg, Sweden},
  publisher = {Association for Computational Linguistics},
  doi       = {10.3115/v1/E14-1056},
  url       = {https://aclanthology.org/E14-1056/}
}

@inproceedings{terragni2021octis,
  title     = {{OCTIS}: comparing and optimizing topic models is simple!},
  author    = {Terragni, Silvia and Fersini, Elisabetta and Galuzzi, Bruno Giovanni and Tropeano, Pietro and Candelieri, Antonio},
  booktitle = {Proceedings of the 16th Conference of the European Chapter of the Association for Computational Linguistics: System Demonstrations},
  month     = apr,
  year      = {2021},
  publisher = {Association for Computational Linguistics},
  url       = {https://aclanthology.org/2021.eacl-demos.31/},
  pages     = {263--270},
  doi       = {10.18653/v1/2021.eacl-demos.31}
}

@inproceedings{wu2024topmost,
  title     = {Towards the {TopMost}: a topic modeling system toolkit},
  author    = {Wu, Xiaobao and Pan, Fengjun and Luu, Anh Tuan},
  editor    = {Cao, Yixin and Feng, Yang and Xiong, Deyi},
  booktitle = {Proceedings of the 62nd Annual Meeting of the Association for Computational Linguistics (Volume 3: System Demonstrations)},
  month     = aug,
  year      = {2024},
  address   = {Bangkok, Thailand},
  publisher = {Association for Computational Linguistics},
  url       = {https://aclanthology.org/2024.acl-demos.4/},
  pages     = {31--41},
  doi       = {10.18653/v1/2024.acl-demos.4}
}

@article{Boyd-Graber:Hu:Mimno-2017,
  title     = {Applications of topic models},
  author    = {Boyd-Graber, Jordan and Hu, Yuening and Mimno, David},
  journal   = {Foundations and Trends in Information Retrieval},
  volume    = {11},
  number    = {2--3},
  pages     = {143--296},
  year      = {2017},
  publisher = {Now Publishers},
  doi       = {10.1561/1500000030},
  url       = {https://www.nowpublishers.com/article/Details/INR-030}
}

@inproceedings{mcauliffe2007supervised,
  title     = {Supervised topic models},
  author    = {McAuliffe, Jon D. and Blei, David M.},
  booktitle = {Advances in Neural Information Processing Systems},
  volume    = {20},
  pages     = {121--128},
  year      = {2007},
  publisher = {Curran Associates, Inc.},
  url       = {https://proceedings.neurips.cc/paper/2007/hash/d56b9fc4b0f1be8871f5e1c40c0067e7-Abstract.html}
}

@incollection{estes2011thematic,
  title     = {Thematic thinking: the apprehension and consequences of thematic relations},
  author    = {Estes, Zachary and Golonka, Sabrina and Jones, Lara L.},
  booktitle = {The Psychology of Learning and Motivation},
  editor    = {Ross, Brian H.},
  series    = {Psychology of Learning and Motivation},
  volume    = {54},
  pages     = {249--294},
  year      = {2011},
  publisher = {Academic Press},
  doi       = {10.1016/B978-0-12-385527-5.00008-5}
}

@article{mohammad2013lexicalcontrast,
  title     = {Computing lexical contrast},
  author    = {Mohammad, Saif M. and Dorr, Bonnie J. and Hirst, Graeme and Turney, Peter D.},
  journal   = {Computational Linguistics},
  volume    = {39},
  number    = {3},
  pages     = {555--590},
  year      = {2013},
  publisher = {MIT Press},
  doi       = {10.1162/COLI_a_00143},
  url       = {https://aclanthology.org/J13-3004/}
}

@inproceedings{ono2015antonym,
  title     = {Word embedding-based antonym detection using thesauri and distributional information},
  author    = {Ono, Masataka and Miwa, Makoto and Sasaki, Yutaka},
  booktitle = {Proceedings of the 2015 Conference of the North American Chapter of the Association for Computational Linguistics: Human Language Technologies},
  pages     = {984--989},
  year      = {2015},
  address   = {Denver, Colorado},
  publisher = {Association for Computational Linguistics},
  url       = {https://aclanthology.org/N15-1100/},
  doi       = {10.3115/v1/N15-1100}
}

@article{budanitsky2006evaluating,
  title   = {Evaluating {WordNet}-based measures of lexical semantic relatedness},
  author  = {Budanitsky, Alexander and Hirst, Graeme},
  journal = {Computational Linguistics},
  volume  = {32},
  number  = {1},
  pages   = {13--47},
  year    = {2006},
  doi     = {10.1162/coli.2006.32.1.13}
}

@article{bruni2014multimodal,
  title   = {Multimodal distributional semantics},
  author  = {Bruni, Elia and Tran, Nam-Khanh and Baroni, Marco},
  journal = {Journal of Artificial Intelligence Research},
  volume  = {49},
  pages   = {1--47},
  year    = {2014},
  doi     = {10.1613/jair.4135},
  url     = {https://jair.org/index.php/jair/article/view/10857}
}

@article{finkelstein2002placing,
  title     = {Placing search in context: the concept revisited},
  author    = {Finkelstein, Lev and Gabrilovich, Evgeniy and Matias, Yossi and Rivlin, Ehud and Solan, Zach and Wolfman, Gadi and Ruppin, Eytan},
  journal   = {ACM Transactions on Information Systems},
  volume    = {20},
  number    = {1},
  pages     = {116--131},
  year      = {2002},
  doi       = {10.1145/503104.503110},
  publisher = {Association for Computing Machinery}
}

@inproceedings{gabrilovich2007computing,
  title     = {Computing semantic relatedness using {Wikipedia}-based explicit semantic analysis},
  author    = {Gabrilovich, Evgeniy and Markovitch, Shaul},
  booktitle = {Proceedings of the 20th International Joint Conference on Artificial Intelligence ({IJCAI})},
  pages     = {1606--1611},
  year      = {2007},
  address   = {Hyderabad, India},
  url       = {https://gabrilovich.com/publications/papers/Gabrilovich2007CSR.pdf}
}

@article{rubenstein1965contextual,
  title     = {Contextual correlates of synonymy},
  author    = {Rubenstein, Herbert and Goodenough, John B.},
  journal   = {Communications of the ACM},
  volume    = {8},
  number    = {10},
  pages     = {627--633},
  year      = {1965},
  doi       = {10.1145/365628.365657},
  publisher = {Association for Computing Machinery}
}

@article{turney2010frequency,
  title   = {From frequency to meaning: vector space models of semantics},
  author  = {Turney, Peter D. and Pantel, Patrick},
  journal = {Journal of Artificial Intelligence Research},
  volume  = {37},
  pages   = {141--188},
  year    = {2010},
  doi     = {10.1613/jair.2934},
  url     = {https://jair.org/index.php/jair/article/view/10640}
}

@inproceedings{devlin2019bert,
  title     = {{BERT}: pre-training of deep bidirectional transformers for language understanding},
  author    = {Devlin, Jacob and Chang, Ming-Wei and Lee, Kenton and Toutanova, Kristina},
  booktitle = {Proceedings of the 2019 Conference of the North American Chapter of the Association for Computational Linguistics: Human Language Technologies, Volume 1 (Long and Short Papers)},
  pages     = {4171--4186},
  year      = {2019},
  address   = {Minneapolis, Minnesota},
  publisher = {Association for Computational Linguistics},
  url       = {https://aclanthology.org/N19-1423/},
  doi       = {10.18653/v1/N19-1423}
}

@inproceedings{gao2021simcse,
  title     = {{SimCSE}: simple contrastive learning of sentence embeddings},
  author    = {Gao, Tianyu and Yao, Xingcheng and Chen, Danqi},
  booktitle = {Proceedings of the 2021 Conference on Empirical Methods in Natural Language Processing},
  pages     = {6894--6910},
  year      = {2021},
  address   = {Online and Punta Cana, Dominican Republic},
  publisher = {Association for Computational Linguistics},
  url       = {https://aclanthology.org/2021.emnlp-main.552/},
  doi       = {10.18653/v1/2021.emnlp-main.552}
}

@article{miller1995wordnet,
  title     = {{WordNet}: a lexical database for English},
  author    = {Miller, George A.},
  journal   = {Communications of the ACM},
  volume    = {38},
  number    = {11},
  pages     = {39--41},
  year      = {1995},
  doi       = {10.1145/219717.219748},
  publisher = {Association for Computing Machinery}
}

@inproceedings{speer2017conceptnet,
  title     = {{ConceptNet} 5.5: an open multilingual graph of general knowledge},
  author    = {Speer, Robyn and Chin, Joshua and Havasi, Catherine},
  booktitle = {Proceedings of the Thirty-First {AAAI} Conference on Artificial Intelligence},
  pages   = {4444--4451},
  year    = {2017},
  publisher = {AAAI Press},
  doi     = {10.1609/aaai.v31i1.11164},
  url     = {https://ojs.aaai.org/index.php/AAAI/article/view/11164}
}

@inproceedings{mou2016tbcnn,
  title     = {Natural language inference by tree-based convolution and heuristic matching},
  author    = {Mou, Lili and Men, Rui and Li, Ge and Xu, Yan and Zhang, Lu and Yan, Rui and Jin, Zhi},
  booktitle = {Proceedings of the 54th Annual Meeting of the Association for Computational Linguistics (Volume 2: Short Papers)},
  pages     = {130--136},
  year      = {2016},
  address   = {Berlin, Germany},
  publisher = {Association for Computational Linguistics},
  url       = {https://aclanthology.org/P16-2022/},
  doi       = {10.18653/v1/P16-2022}
}

@inproceedings{conneau2017infersent,
  title     = {Supervised learning of universal sentence representations from natural language inference data},
  author    = {Conneau, Alexis and Kiela, Douwe and Schwenk, Holger and Barrault, Lo{\"i}c and Bordes, Antoine},
  booktitle = {Proceedings of the 2017 Conference on Empirical Methods in Natural Language Processing},
  pages     = {670--680},
  year      = {2017},
  address   = {Copenhagen, Denmark},
  publisher = {Association for Computational Linguistics},
  url       = {https://aclanthology.org/D17-1070/},
  doi       = {10.18653/v1/D17-1070}
}

@misc{lewis1997reuters21578,
  author = {Lewis, David D.},
  title  = {Reuters-21578 text categorization test collection},
  year   = {1997},
  url    = {https://kdd.ics.uci.edu/databases/reuters21578/reuters21578.html}
}

@inproceedings{pennington2014glove,
  title     = {{GloVe}: global vectors for word representation},
  author    = {Pennington, Jeffrey and Socher, Richard and Manning, Christopher D.},
  booktitle = {Proceedings of the 2014 Conference on Empirical Methods in Natural Language Processing ({EMNLP})},
  pages     = {1532--1543},
  year      = {2014},
  address   = {Doha, Qatar},
  publisher = {Association for Computational Linguistics},
  url       = {https://aclanthology.org/D14-1162/},
  doi       = {10.3115/v1/D14-1162}
}

@article{wu2024survey,
  title     = {A survey on neural topic models: methods, applications, and challenges},
  author    = {Wu, Xiaobao and Nguyen, Thong and Luu, Anh Tuan},
  journal   = {Artificial Intelligence Review},
  volume    = {57},
  number    = {2},
  pages     = {18},
  year      = {2024},
  publisher = {Springer},
  doi       = {10.1007/s10462-023-10661-7},
  url       = {https://doi.org/10.1007/s10462-023-10661-7}
}

@inproceedings{lafferty2005correlated,
  title     = {Correlated topic models},
  author    = {Lafferty, John and Blei, David},
  booktitle = {Advances in Neural Information Processing Systems},
  volume    = {18},
  pages     = {147--154},
  year      = {2005},
  publisher = {MIT Press}
}

@inproceedings{rahimi2024contextualized,
  title     = {Contextualized topic coherence metrics},
  author    = {Rahimi, Hamed and Mimno, David and Hoover, Jacob and Naacke, Hubert and Constantin, Camelia and Amann, Bernd},
  booktitle = {Findings of the Association for Computational Linguistics: {EACL} 2024},
  pages     = {1760--1773},
  year      = {2024},
  address   = {St. Julian's, Malta},
  publisher = {Association for Computational Linguistics},
  url       = {https://aclanthology.org/2024.findings-eacl.123/},
  doi       = {10.18653/v1/2024.findings-eacl.123}
}

@inproceedings{gupta2021obtaining,
  title     = {Obtaining better static word embeddings using contextual embedding models},
  author    = {Gupta, Prakhar and Jaggi, Martin},
  booktitle = {Proceedings of the 59th Annual Meeting of the Association for Computational Linguistics and the 11th International Joint Conference on Natural Language Processing (Volume 1: Long Papers)},
  pages     = {5241--5253},
  year      = {2021},
  address   = {Online},
  publisher = {Association for Computational Linguistics},
  url       = {https://aclanthology.org/2021.acl-long.408/},
  doi       = {10.18653/v1/2021.acl-long.408}
}

@inproceedings{resnik1995using,
  title     = {Using information content to evaluate semantic similarity in a taxonomy},
  author    = {Resnik, Philip},
  booktitle = {Proceedings of the 14th International Joint Conference on Artificial Intelligence ({IJCAI})},
  pages   = {448--453},
  year    = {1995},
  url     = {https://www.ijcai.org/Proceedings/95-1/Papers/059.pdf}
}

@techreport{deepseekai2026v4,
  title       = {{DeepSeek-V4}: Towards Highly Efficient Million-Token Context Intelligence},
  author      = {{DeepSeek-AI}},
  institution = {DeepSeek-AI},
  type        = {Technical Report},
  year        = {2026},
  url         = {https://arxiv.org/abs/2606.19348},
  doi         = {10.48550/arXiv.2606.19348},
  note        = {Annotator: API model \texttt{deepseek-v4-flash}, accessed 2026-05-24.}
}

@techreport{openai2025gpt41,
  title       = {Introducing {GPT-4.1} in the {API}},
  author      = {{OpenAI}},
  institution = {OpenAI},
  type        = {Technical Report},
  year        = {2025},
  url         = {https://openai.com/index/gpt-4-1/},
  note        = {Annotator: API model \texttt{gpt-4.1-mini}, accessed 2026-05-24.}
}

@techreport{qwen2025qwenturbo,
  title       = {{Model Studio} Update of {Qwen-Plus/Turbo}},
  author      = {{Alibaba Cloud Model Studio}},
  institution = {Alibaba Cloud},
  type        = {Service Update},
  year        = {2025},
  url         = {https://www.alibabacloud.com/en/notice/model_studio_update_of_qwenplusturbo_4f8},
  note        = {Annotator: API model \texttt{qwen-turbo-latest}, resolving to the \texttt{qwen-turbo-2025-04-28} snapshot; accessed 2026-05-24.}
}

@techreport{xai2026grok420,
  title       = {{Grok} 4.20 (Non-Reasoning)},
  author      = {{xAI}},
  institution = {xAI},
  type        = {Model Card},
  year        = {2026},
  url         = {https://docs.x.ai/developers/models/grok-4.20-0309-non-reasoning},
  note        = {Ablation-only annotator: API model \texttt{grok-4.20-0309-non-reasoning}, accessed 2026-05-21.}
}

@techreport{anthropic2025claude45,
  title       = {{Claude} Haiku 4.5 System Card},
  author      = {{Anthropic}},
  institution = {Anthropic},
  type        = {Technical Report},
  year        = {2025},
  month       = oct,
  url         = {https://assets.anthropic.com/m/99128ddd009bdcb/original/Claude-Haiku-4-5-System-Card.pdf},
  note        = {Validation-only annotator: API model \texttt{claude-haiku-4-5}, accessed 2026-05-24.}
}

@inproceedings{ranjan2026oneword,
  title     = {One Word Is Not Enough: Simple Prompts Improve Word Embeddings},
  author    = {Ranjan, Rajeev},
  booktitle = {Proceedings of the 15th Joint Conference on Lexical and Computational Semantics (*{SEM} 2026)},
  month     = jul,
  year      = {2026},
  address   = {San Diego, California, United States},
  publisher = {Association for Computational Linguistics},
  pages     = {464--473},
  url       = {https://aclanthology.org/2026.starsem-conference.32/},
  doi       = {10.18653/v1/2026.starsem-conference.32}
}

@misc{openai2024embedding,
  title  = {New embedding models and {API} updates},
  author = {{OpenAI}},
  year   = {2024},
  url    = {https://openai.com/index/new-embedding-models-and-api-updates/},
  note   = {Accessed: 2026-05-23}
}

@article{liu2019roberta,
  title   = {{RoBERTa}: A Robustly Optimized {BERT} Pretraining Approach},
  author  = {Liu, Yinhan and Ott, Myle and Goyal, Naman and Du, Jingfei and Joshi, Mandar and Chen, Danqi and Levy, Omer and Lewis, Mike and Zettlemoyer, Luke and Stoyanov, Veselin},
  journal = {arXiv preprint arXiv:1907.11692},
  year    = {2019},
  url     = {https://arxiv.org/abs/1907.11692}
}

@inproceedings{lim2015cntm_m10,
  title     = {Bibliographic analysis with the citation network topic model},
  author    = {Lim, Kar Wai and Buntine, Wray},
  booktitle = {Proceedings of the Asian Conference on Machine Learning ({ACML})},
  series    = {Proceedings of Machine Learning Research},
  volume    = {39},
  pages     = {142--158},
  year      = {2015},
  publisher = {PMLR},
}

@inproceedings{ley2002dblp,
  title     = {The {DBLP} computer science bibliography: evolution, research issues, perspectives},
  author    = {Ley, Michael},
  booktitle = {String Processing and Information Retrieval ({SPIRE})},
  series    = {Lecture Notes in Computer Science},
  volume    = {2476},
  pages     = {1--10},
  year      = {2002},
  publisher = {Springer},
  doi       = {10.1007/3-540-45735-6_1},
  url       = {https://doi.org/10.1007/3-540-45735-6_1}
}

@inproceedings{bird2008aclarc,
  title     = {The {ACL} anthology reference corpus: a reference dataset for bibliographic research in computational linguistics},
  author    = {Bird, Steven and Dale, Robert and Dorr, Bonnie and Gibson, Bryan and Joseph, Mark and Kan, Min-Yen and Lee, Dongwon and Powley, Brett and Radev, Dragomir and Tan, Yee Fan},
  booktitle = {Proceedings of the Sixth International Conference on Language Resources and Evaluation ({LREC}'08)},
  pages   = {1755--1759},
  year    = {2008},
  url     = {http://www.lrec-conf.org/proceedings/lrec2008/pdf/445_paper.pdf}
}

@inproceedings{greene2006bbc,
  title     = {Practical solutions to the problem of diagonal dominance in kernel document clustering},
  author    = {Greene, Derek and Cunningham, P{\'a}draig},
  booktitle = {Proceedings of the 23rd International Conference on Machine Learning ({ICML})},
  pages     = {377--384},
  year      = {2006},
  publisher = {Association for Computing Machinery},
  doi       = {10.1145/1143844.1143892},
  url       = {https://doi.org/10.1145/1143844.1143892}
}

@inproceedings{lang1995newsweeder,
  title     = {Newsweeder: learning to filter netnews},
  author    = {Lang, Ken},
  booktitle = {Proceedings of the Twelfth International Conference on Machine Learning ({ICML})},
  pages     = {331--339},
  year      = {1995},
  publisher = {Morgan Kaufmann},
  doi       = {10.1016/B978-1-55860-377-6.50048-7},
  url       = {https://doi.org/10.1016/B978-1-55860-377-6.50048-7}
}

@book{kendall1948rank,
  title     = {Rank correlation methods},
  author    = {Kendall, Maurice G.},
  year      = {1948},
  publisher = {Charles Griffin \& Company},
  address   = {London}
}

@inproceedings{tran2024enhancing,
  title     = {Enhancing knowledge retrieval with topic modeling for knowledge-grounded dialogue},
  author    = {Tran, Nhat and Litman, Diane},
  booktitle = {Proceedings of the 2024 Joint International Conference on Computational Linguistics, Language Resources and Evaluation ({LREC}-{COLING} 2024)},
  pages     = {5986--5995},
  month     = may,
  year      = {2024},
  address   = {Torino, Italia},
  publisher = {ELRA and ICCL},
  url       = {https://aclanthology.org/2024.lrec-main.530/}
}

@inproceedings{lewis2020retrieval,
  title     = {Retrieval-augmented generation for knowledge-intensive {NLP} tasks},
  author    = {Lewis, Patrick and Perez, Ethan and Piktus, Aleksandra and Petroni, Fabio and Karpukhin, Vladimir and Goyal, Naman and K{\"u}ttler, Heinrich and Lewis, Mike and Yih, Wen-tau and Rockt{\"a}schel, Tim and Riedel, Sebastian and Kiela, Douwe},
  booktitle = {Advances in Neural Information Processing Systems},
  volume    = {33},
  pages     = {9459--9474},
  year      = {2020},
  publisher = {Curran Associates, Inc.},
  url       = {https://papers.nips.cc/paper_files/paper/2020/hash/6b493230205f780e1bc26945df7481e5-Abstract.html}
}

@inproceedings{wu2024kpeval,
  title     = {Kpeval: towards fine-grained semantic-based keyphrase evaluation},
  author    = {Wu, Di and Yin, Da and Chang, Kai-Wei},
  booktitle = {Findings of the Association for Computational Linguistics: {ACL} 2024},
  pages     = {1959--1981},
  year      = {2024},
  address   = {Bangkok, Thailand},
  publisher = {Association for Computational Linguistics},
  url       = {https://aclanthology.org/2024.findings-acl.117/},
  doi       = {10.18653/v1/2024.findings-acl.117}
}

@article{bojanowski2017enriching,
  title   = {Enriching word vectors with subword information},
  author  = {Bojanowski, Piotr and Grave, Edouard and Joulin, Armand and Mikolov, Tomas},
  journal = {Transactions of the Association for Computational Linguistics},
  volume  = {5},
  pages   = {135--146},
  year    = {2017},
  doi     = {10.1162/tacl_a_00051},
  url     = {https://aclanthology.org/Q17-1010/}
}

@article{misra2022newscategory,
  title   = {News Category Dataset},
  author  = {Misra, Rishabh},
  journal = {arXiv preprint arXiv:2209.11429},
  year    = {2022},
  doi     = {10.48550/arXiv.2209.11429},
  url     = {https://arxiv.org/abs/2209.11429}
}

@article{petukhova2023mnds,
  title     = {{MN-DS}: A Multilabeled News Dataset for News Articles Hierarchical Classification},
  author    = {Petukhova, Alina and Fachada, Nuno},
  journal   = {Data},
  volume    = {8},
  number    = {5},
  pages     = {74},
  year      = {2023},
  publisher = {MDPI},
  doi       = {10.3390/data8050074},
  url       = {https://www.mdpi.com/2306-5729/8/5/74}
}

@inproceedings{wu1994verb,
  title     = {Verb Semantics and Lexical Selection},
  author    = {Wu, Zhibiao and Palmer, Martha},
  booktitle = {Proceedings of the 32nd Annual Meeting of the Association for Computational Linguistics},
  pages     = {133--138},
  year      = {1994},
  address   = {Las Cruces, New Mexico, USA},
  publisher = {Association for Computational Linguistics},
  url       = {https://aclanthology.org/P94-1019/},
  doi       = {10.3115/981732.981751}
}

@article{clement2019arxiv,
  title   = {On the Use of {ArXiv} as a Dataset},
  author  = {Clement, Colin B. and Bierbaum, Matthew and O'Keeffe, Kevin P. and Alemi, Alexander A.},
  journal = {arXiv preprint arXiv:1905.00075},
  year    = {2019},
  doi     = {10.48550/arXiv.1905.00075},
  url     = {https://arxiv.org/abs/1905.00075}
}

@inproceedings{schick2021generating,
  title     = {Generating Datasets with Pretrained Language Models},
  author    = {Schick, Timo and Sch{\"u}tze, Hinrich},
  booktitle = {Proceedings of the 2021 Conference on Empirical Methods in Natural Language Processing},
  pages     = {6943--6951},
  year      = {2021},
  address   = {Online and Punta Cana, Dominican Republic},
  publisher = {Association for Computational Linguistics},
  url       = {https://aclanthology.org/2021.emnlp-main.555/},
  doi       = {10.18653/v1/2021.emnlp-main.555}
}

@inproceedings{west2022symbolic,
  title     = {Symbolic Knowledge Distillation: from General Language Models to Commonsense Models},
  author    = {West, Peter and Bhagavatula, Chandra and Hessel, Jack and Hwang, Jena D. and Jiang, Liwei and Le Bras, Ronan and Lu, Ximing and Welleck, Sean and Choi, Yejin},
  booktitle = {Proceedings of the 2022 Conference of the North American Chapter of the Association for Computational Linguistics: Human Language Technologies},
  pages     = {4602--4625},
  year      = {2022},
  address   = {Seattle, United States},
  publisher = {Association for Computational Linguistics},
  url       = {https://aclanthology.org/2022.naacl-main.341/},
  doi       = {10.18653/v1/2022.naacl-main.341}
}

@article{blei2012probabilistic,
  title   = {Probabilistic Topic Models},
  author  = {Blei, David M.},
  journal = {Communications of the ACM},
  volume  = {55},
  number  = {4},
  pages   = {77--84},
  year    = {2012},
  doi     = {10.1145/2133806.2133826},
  url     = {https://dl.acm.org/doi/10.1145/2133806.2133826}
}

@article{grimmer2013text,
  title   = {Text as Data: The Promise and Pitfalls of Automatic Content Analysis Methods for Political Texts},
  author  = {Grimmer, Justin and Stewart, Brandon M.},
  journal = {Political Analysis},
  volume  = {21},
  number  = {3},
  pages   = {267--297},
  year    = {2013},
  doi     = {10.1093/pan/mps028}
}

@inproceedings{pham2024topicgpt,
  title     = {{TopicGPT}: A Prompt-based Topic Modeling Framework},
  author    = {Pham, Chau Minh and Hoyle, Alexander and Sun, Simeng and Resnik, Philip and Iyyer, Mohit},
  booktitle = {Proceedings of the 2024 Conference of the North American Chapter of the Association for Computational Linguistics: Human Language Technologies (Volume 1: Long Papers)},
  pages     = {2956--2984},
  year      = {2024},
  address   = {Mexico City, Mexico},
  publisher = {Association for Computational Linguistics},
  url       = {https://aclanthology.org/2024.naacl-long.164/},
  doi       = {10.18653/v1/2024.naacl-long.164}
}

@inproceedings{mu2024large,
  title     = {Large Language Models Offer an Alternative to the Traditional Approach of Topic Modelling},
  author    = {Mu, Yida and Dong, Chun and Bontcheva, Kalina and Song, Xingyi},
  booktitle = {Proceedings of the 2024 Joint International Conference on Computational Linguistics, Language Resources and Evaluation (LREC-COLING 2024)},
  pages     = {10160--10171},
  year      = {2024},
  address   = {Torino, Italia},
  publisher = {ELRA and ICCL},
  url       = {https://aclanthology.org/2024.lrec-main.887/}
}

@article{hansen2018transparency,
  title     = {Transparency and Deliberation Within the {FOMC}: A Computational Linguistics Approach},
  author    = {Hansen, Stephen and McMahon, Michael and Prat, Andrea},
  journal   = {The Quarterly Journal of Economics},
  volume    = {133},
  number    = {2},
  pages     = {801--870},
  year      = {2018},
  doi       = {10.1093/qje/qjx045}
}

@article{barbera2019who,
  title     = {Who Leads? {W}ho Follows? {M}easuring Issue Attention and Agenda Setting by Legislators and the Mass Public Using Social Media Data},
  author    = {Barber{\'a}, Pablo and Casas, Andreu and Nagler, Jonathan and Egan, Patrick J. and Bonneau, Richard and Jost, John T. and Tucker, Joshua A.},
  journal   = {American Political Science Review},
  volume    = {113},
  number    = {4},
  pages     = {883--901},
  year      = {2019},
  doi       = {10.1017/S0003055419000352}
}

@article{liu2021tracing,
  title     = {Tracing the Pace of {COVID}-19 Research: Topic Modeling and Evolution},
  author    = {Liu, Jiaying and Nie, Hansong and Li, Shihao and Chen, Xiangtai and Cao, Huazhu and Ren, Jing and Lee, Ivan and Xia, Feng},
  journal   = {Big Data Research},
  volume    = {25},
  pages     = {100236},
  year      = {2021},
  doi       = {10.1016/j.bdr.2021.100236}
}

@article{egger2022topic,
  title     = {A Topic Modeling Comparison Between {LDA}, {NMF}, {Top2Vec}, and {BERTopic} to Demystify {Twitter} Posts},
  author    = {Egger, Roman and Yu, Joanne},
  journal   = {Frontiers in Sociology},
  volume    = {7},
  pages     = {886498},
  year      = {2022},
  doi       = {10.3389/fsoc.2022.886498}
}

@article{suter2025when,
  title     = {When Politicians Talk {AI}: Issue-Frames in Parliamentary Debates Before and After {ChatGPT}},
  author    = {Suter, Viktor and Ma, Charles and P{\"o}hlmann, Gina and Meckel, Miriam},
  journal   = {Policy \& Internet},
  volume    = {17},
  number    = {3},
  pages     = {e70010},
  year      = {2025},
  doi       = {10.1002/poi3.70010}
}

@article{tangherlini2013trawling,
  title     = {Trawling in the Sea of the Great Unread: Sub-corpus Topic Modeling and Humanities Research},
  author    = {Tangherlini, Timothy R. and Leonard, Peter},
  journal   = {Poetics},
  volume    = {41},
  number    = {6},
  pages     = {725--749},
  year      = {2013},
  doi       = {10.1016/j.poetic.2013.08.002}
}

\clearpage

\appendix

\renewcommand{\topfraction}{0.95}
\renewcommand{\bottomfraction}{0.7}
\renewcommand{\textfraction}{0.07}
\renewcommand{\floatpagefraction}{0.6}
\renewcommand{\dbltopfraction}{0.95}
\renewcommand{\dblfloatpagefraction}{0.5}
\setcounter{topnumber}{3}
\setcounter{bottomnumber}{2}
\setcounter{totalnumber}{5}
\setcounter{dbltopnumber}{3}
\makeatletter
\setlength{\@fptop}{0pt}
\setlength{\@dblfptop}{0pt}
\makeatother

\setlength{\intextsep}{4pt plus 2pt minus 2pt}
\setlength{\floatsep}{4pt plus 2pt minus 2pt}
\setlength{\textfloatsep}{4pt plus 2pt minus 2pt}
\setlength{\abovecaptionskip}{3pt}
\setlength{\belowcaptionskip}{0pt}

\section{Complementary results}

    \subsection{Inconsistent Scores in Existing Datasets}
    \label{app:inconsistent}

    Table~\ref{tab:word_pairs} makes the conflation noted in the introduction concrete. Because each resource operationalizes only one dimension, the same pair can receive near-opposite normalized scores across datasets. \emph{Breakfast}--\emph{dinner} is rated 0.941 by TxThmNorms but 0.149 by SimLex-999: the two meals are strong co-hyponyms of a shared category, which the taxonomic-similarity annotation rewards, yet SimLex-999 penalizes them for not being substitutable. \emph{Camel}--\emph{desert} shows the same split on the relatedness axis, at 0.903 in TxThmNorms against 0.580 in MEN, because a thematic role relation is scored more conservatively under MEN's broader distributional notion of relatedness. Neither score is wrong on its own terms; the disagreement is a direct consequence of collapsing two distinct dimensions onto one scale, and it is why we do not train on a pooled concatenation of these resources.

    \begin{table*}[t]
    \centering
    \caption{Conflicting scores across different datasets (normalized).}
    \label{tab:word_pairs}
    \small
    \begin{tabular}{llrrlr}
    \toprule
    Word Pair & Score Type & Score 1 & Score 2 & Datasource 1 & Datasource 2 \\
    \midrule
    breakfast / dinner & similarity & 0.941 & 0.149 & TxThmNorms & SimLex-999 \\
    camel / desert & relatedness & 0.903 & 0.580 & TxThmNorms & MEN \\
    \end{tabular}
    \end{table*}

    \subsection{BERT-KT vs.\ LDA Topic Examples}
    \label{app:bertkt}
    
    Table~\ref{tab:lda_bertkt_example_topics} illustrates the contrast. BERT-KT topics group near-synonyms and antonyms (e.g.\ variants of ``rise,'' or institutional role titles), reflecting embedding-space proximity rather than contextual interaction. LDA topics instead comprise contextually related words with complementary semantic roles, capturing distributional co-occurrence rather than categorical similarity.
    
    \begin{table}[htbp]
    \centering
    \caption{Representative topics generated by LDA and BERT-KT on Reuters~\citep{sia2020tired}. Each topic is shown using its top 5 words.}
    \label{tab:lda_bertkt_example_topics}
    \small
    \setlength{\tabcolsep}{4pt}
    \renewcommand{\arraystretch}{1.15}
    \begin{tabular}{@{}c p{0.36\linewidth} p{0.36\linewidth}@{}}
    \toprule
     & \textbf{LDA (Reuters)} & \textbf{BERT-KT (Reuters)} \\
    \midrule

    \textbf{Topic 1} &
    dollar, rate, exchange, currency, market &
    rise, increase, fall, decline, drop \\

    \textbf{Topic 2} &
    growth, government, economic, economy, inflation &
    president, chairman, minister, secretary, chief \\

    \textbf{Topic 3} &
    gold, reserves, production, ounces, mine &
    make, continue, reduce, support, remain \\

    \bottomrule
    \end{tabular}
    \end{table}

    \subsection{Downstream Task Performance Tables}
    \label{app:task_tables}
    \label{app:taskc_per_corpus}
    Tables~\ref{tab:taskAB} and~\ref{tab:taskC} report full results for the three downstream tasks: Event Monitoring, Category Retrieval, and Synonym-Aware Retrieval.

    \begin{table*}[t]
    \centering
    \caption{Event Monitoring and Category Retrieval: model performance on the Reuters corpus.}
    \label{tab:taskAB}
    \small
    \begin{tabular}{ll rr r rrr rr}
\toprule
& & \multicolumn{2}{c}{Task A} & Task B & \multicolumn{5}{c}{Metric Scores} \\
\cmidrule(lr){3-4} \cmidrule(lr){5-5} \cmidrule(lr){6-10}
Model & Nature & Hit@3 & Recall@3 & SCR & Sim. & Rel. & Gap & Coh. & Div. \\
\midrule
lda          & Non-PLM  & 0.585 & 0.492 & 0.688 & 0.246 & 0.582 & $-$0.336 & 0.510 & 0.660 \\
bertopic     & PLM-Aug. & 0.573 & 0.482 & 0.679 & 0.228 & 0.558 & $-$0.330 & 0.612 & 0.824 \\
nmf          & Non-PLM  & 0.526 & 0.456 & 0.642 & 0.251 & 0.572 & $-$0.321 & 0.547 & 0.618 \\
neurallda    & Non-PLM  & 0.523 & 0.441 & 0.624 & 0.224 & 0.537 & $-$0.313 & 0.417 & 0.998 \\
prodlda      & Non-PLM  & 0.520 & 0.445 & 0.662 & 0.233 & 0.532 & $-$0.299 & 0.629 & 0.868 \\
ctm          & PLM-Aug. & 0.493 & 0.419 & 0.650 & 0.242 & 0.545 & $-$0.303 & 0.633 & 0.875 \\
bertkt       & PLM-Aug. & 0.412 & 0.353 & 0.574 & 0.439 & 0.595 & $-$0.156 & 0.495 & 0.965 \\
lsi          & Non-PLM  & 0.397 & 0.348 & 0.577 & 0.209 & 0.573 & $-$0.364 & 0.373 & 0.370 \\
fastopic     & PLM-Aug. & 0.329 & 0.286 & 0.615 & 0.202 & 0.414 & $-$0.212 & ---   & 1.00  \\
tntm         & PLM-Aug. & 0.247 & 0.204 & 0.330 & 0.279 & 0.480 & $-$0.201 & 0.471 & 1.00  \\
thompson\_mimno & PLM-Aug. & 0.159 & 0.147 & 0.374 & 0.347 & 0.569 & $-$0.222 & ---   & 1.00  \\
random       & Baseline & 0.0904 & 0.0784 & 0.302 & 0.110 & 0.322 & $-$0.212 & ---   & 1.00  \\
\bottomrule
\end{tabular}

    \end{table*}
    
    \begin{table*}[t]
    \centering
    \caption{Synonym-Aware Retrieval: Jaccard@10 across the nine corpora. ``---'' indicates the model was excluded for that corpus (genuine topic collapse) or not run.}
    \label{tab:taskC}
    \small
    \begin{tabular}{ll ccccccccc}
\toprule
Model & Type & Reut. & 20NG & BBC & DBLP & ACL & M10 & NewsC & MN-DS & arXiv \\
\midrule
\multicolumn{11}{l}{\textit{PLM Augmented}} \\
TNTM    & PLM Aug.     & \textbf{0.94} & ---  & \textbf{0.96} & ---  & ---  & \textbf{0.70} & \textbf{0.94} & \textbf{0.98} & \textbf{0.97} \\
BERT-KT & PLM Aug.     & 0.76 & \textbf{0.48} & 0.40 & \textbf{0.60} & \textbf{0.68} & 0.54 & 0.74 & 0.74 & 0.56 \\
T\&M    & PLM Aug.     & 0.45 & 0.25 & 0.38 & 0.39 & ---  & 0.33 & ---  & ---  & ---  \\
\midrule
\multicolumn{11}{l}{\textit{Non-PLM Augmented}} \\
LDA     & Non-PLM Aug. & 0.23 & 0.16 & 0.12 & 0.06 & 0.20 & 0.13 & 0.04 & 0.12 & 0.17 \\
ProdLDA & Non-PLM Aug. & 0.15 & 0.08 & 0.14 & 0.14 & 0.19 & 0.10 & 0.27 & 0.26 & 0.56 \\
\bottomrule
\end{tabular}

    \end{table*}
    
    \begin{figure}[htbp]
        \centering
        \includegraphics[width=\linewidth]{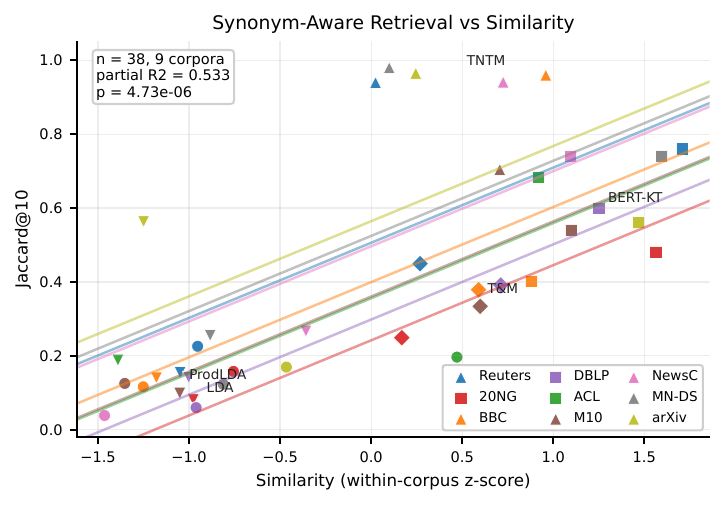}
        \caption{Synonym-Aware Retrieval regression with corpus fixed effects: similarity score predicts synonym-aware retrieval consistency (Jaccard@10) across the nine corpora (partial $R^2 = 0.53$, $p < 0.001$). Parallel lines show per-corpus intercepts with a shared slope.}
        \label{fig:taskC}
    \end{figure}

    \subsection{Synonym-Aware Retrieval: Word-Pair Inventory}
    \label{app:synonym_pairs}

    Synonym-Aware Retrieval uses 50 single-token synonym pairs per corpus, validated so that both members appear with non-zero frequency in the corpus's preprocessed vocabulary. Two shared lists cover most corpora: a \emph{technical/scientific} list (Table~\ref{tab:syn_pairs_tech}; 20NewsGroups, ACL, DBLP, M10) and a \emph{business/news} list (Table~\ref{tab:syn_pairs_news}; Reuters, BBC News). The three larger corpora use per-corpus variants of one of these lists: arXiv swaps four lemmas relative to the technical list; MN-DS and News Category each substitute roughly fifteen entries from the business/news list with domain-fit alternatives (e.g.\ \emph{confirm/verify}, \emph{eliminate/remove}, \emph{frequent/common}) when an original lemma is absent from the corpus vocabulary. The variant YAMLs are released alongside the code.

    \begin{table*}[t]
    \centering
    \caption{Technical/scientific synonym list (50 pairs). Shared by 20NewsGroups, ACL, DBLP, and M10; arXiv uses a four-lemma variant of this list.}
    \label{tab:syn_pairs_tech}
    \scriptsize
    \setlength{\tabcolsep}{4pt}
    \renewcommand{\arraystretch}{1.05}
    \begin{tabular}{l l l l}
    \toprule
    approach/method            & evaluate/assess              & evaluation/assessment      & identify/detect           \\
    identification/detection   & discover/find                & analysis/study             & model/framework           \\
    method/technique           & procedure/process            & improve/enhance            & develop/design            \\
    combine/integrate          & predict/estimate             & prediction/estimation      & approximate/estimate      \\
    approximation/estimation   & accuracy/precision           & robust/stable              & stability/consistency     \\
    verification/validation    & goal/objective               & goal/target                & objective/target          \\
    task/work                  & handle/manage                & assist/support             & build/construct           \\
    construction/building      & create/generate              & creation/generation        & computation/analysis      \\
    infer/derive               & inference/interpretation     & measure/assessment         & compare/contrast          \\
    comparison/contrast        & difference/variation         & variation/variance         & vary/change               \\
    choice/selection           & class/category               & component/element          & architecture/structure    \\
    design/architecture        & implement/apply              & implementation/application & develop/build             \\
    development/construction   & learning/training            &                            &                           \\
    \bottomrule
    \end{tabular}
    \end{table*}

    \begin{table*}[t]
    \centering
    \caption{Business/news synonym list (50 pairs). Shared by Reuters and BBC News; MN-DS and News Category use variants of this list with $\sim$15 substitutions each for lemmas absent from the corpus vocabulary.}
    \label{tab:syn_pairs_news}
    \scriptsize
    \setlength{\tabcolsep}{4pt}
    \renewcommand{\arraystretch}{1.05}
    \begin{tabular}{l l l l}
    \toprule
    increase/rise        & decline/fall          & reduce/lower         & stop/halt              \\
    purchase/acquire     & agreement/accord      & company/firm         & stock/share            \\
    lawsuit/suit         & announce/declare      & begin/start          & benefit/advantage      \\
    change/alter         & choose/select         & close/shut           & complete/finish        \\
    concern/worry        & create/build          & deal/agreement       & delay/postpone         \\
    demand/request       & deny/reject           & dismiss/reject       & estimate/forecast      \\
    expand/grow          & fast/quick            & final/ultimate       & goal/objective         \\
    happen/occur         & important/significant & improve/enhance      & income/revenue         \\
    lead/guide           & limit/restrict        & main/primary         & major/significant      \\
    money/cash           & obtain/acquire        & offer/propose        & opinion/view           \\
    oppose/resist        & order/request         & plan/scheme          & prevent/avoid          \\
    profit/gain          & promise/pledge        & provide/supply       & quickly/rapidly        \\
    raise/increase       & result/outcome        &                      &                        \\
    \bottomrule
    \end{tabular}
    \end{table*}

\subsection{Downstream Regression: Specification}
    \label{app:regression_method}

    All downstream regressions use a single pooled cross-corpus fixed-effects
    specification, described here for reproducibility.

    \paragraph{Panel and fixed effects.} Each observation is one (model, corpus)
    pair: a topic model's mean intrinsic score on a corpus paired with its
    downstream performance on the same corpus. The two retrieval tasks pool four labeled
    corpora ($n{=}46$) and Synonym-Aware Retrieval pools nine ($n{=}38$). Every regression
    includes a corpus fixed effect $C(\text{corpus})$---a per-corpus
    intercept---so that the predictor's slope is identified only by
    within-corpus, cross-model variation; baseline differences between corpora
    (label granularity, document length, metric scale) are absorbed rather than
    confounded with the predictor.

    \paragraph{Standardisation.} Each predictor is $z$-scored within each corpus
    (population mean and standard deviation over the models evaluated on that
    corpus) before pooling, so a coefficient is a per-standard-deviation effect
    and is comparable across predictors that live on very different raw scales
    (e.g.\ UMass coherence vs.\ inverted RBO). For Event Monitoring and Category Retrieval the outcome is
    likewise $z$-scored within corpus, since the four corpora use different
    performance metrics; Synonym-Aware Retrieval uses the raw Jaccard@10, already a bounded,
    corpus-comparable quantity.

    \paragraph{Single-predictor partial $R^2$.} Table~\ref{tab:single_factor}
    enters one predictor at a time and reports
    its \emph{partial} $R^2$---the share of outcome variance it explains
    \emph{beyond} the corpus fixed effects,
    $R^2_{\text{partial}} = (\text{RSS}_{\text{red}} - \text{RSS}_{\text{full}})
    / \text{RSS}_{\text{red}}$, where the reduced model is
    \texttt{outcome}\,$\sim$\,$C(\text{corpus})$ and the full model adds the
    predictor. Unlike the raw model $R^2$, the partial $R^2$ is not inflated by
    the fixed effects and stays comparable regardless of how many corpora a
    given predictor can be scored on.

    \paragraph{Joint similarity--relatedness model.} The joint specification
    enters similarity and relatedness \emph{together},
    \texttt{outcome}\,$\sim$\,\texttt{sim}\,$+$\,\texttt{rel}\,$+$\,$C(\text{corpus})$,
    both $z$-scored within corpus, so each coefficient is the partial effect of
    one axis holding the other fixed (Table~\ref{tab:joint_simrel}). Because the
    two raw scores are strongly positively correlated (per-corpus Pearson
    $r \approx 0.73$--$0.95$, pooled within-corpus $r \approx 0.84$),
    only this joint fit separates their genuine opposing contributions from the
    shared topic-quality component; the single-predictor coefficient of the
    mismatched axis is positively biased by suppression.

    \begin{table}[H]
    \centering
    \caption{Joint cross-corpus fixed-effects regression
    \texttt{outcome}\,$\sim$\,\texttt{sim}\,$+$\,\texttt{rel}\,$+$\,$C$(corpus),
    both predictors $z$-scored within corpus. Each coefficient is the partial
    effect of one axis controlling for the other; $R^2$ is the full-model value
    (includes the corpus fixed effects). \textbf{Bold}: $p<.001$.}
    \label{tab:joint_simrel}
    \small
    \setlength{\tabcolsep}{5pt}
    \begin{tabular}{@{}lcccc@{}}
    \toprule
    \textbf{Task} & $n$ & $\beta_{\text{sim}}$ & $\beta_{\text{rel}}$ & $R^2$ \\
    \midrule
    Event Monitoring   & 46 & $\mathbf{-1.01}$ & $\mathbf{+1.31}$ & .58 \\
    Category Retrieval & 46 & $\mathbf{-0.88}$ & $\mathbf{+1.37}$ & .68 \\
    Synonym-Aware Retrieval  & 38 & $\mathbf{+0.34}$ & $\mathbf{-0.19}$ & .79 \\
    \bottomrule
    \end{tabular}
    \end{table}

    \subsection{Atlas Visualizations by Corpus}
    \label{app:atlas_viz}

\begin{figure*}[tp]
        \centering
        \includegraphics[width=0.94\textwidth]{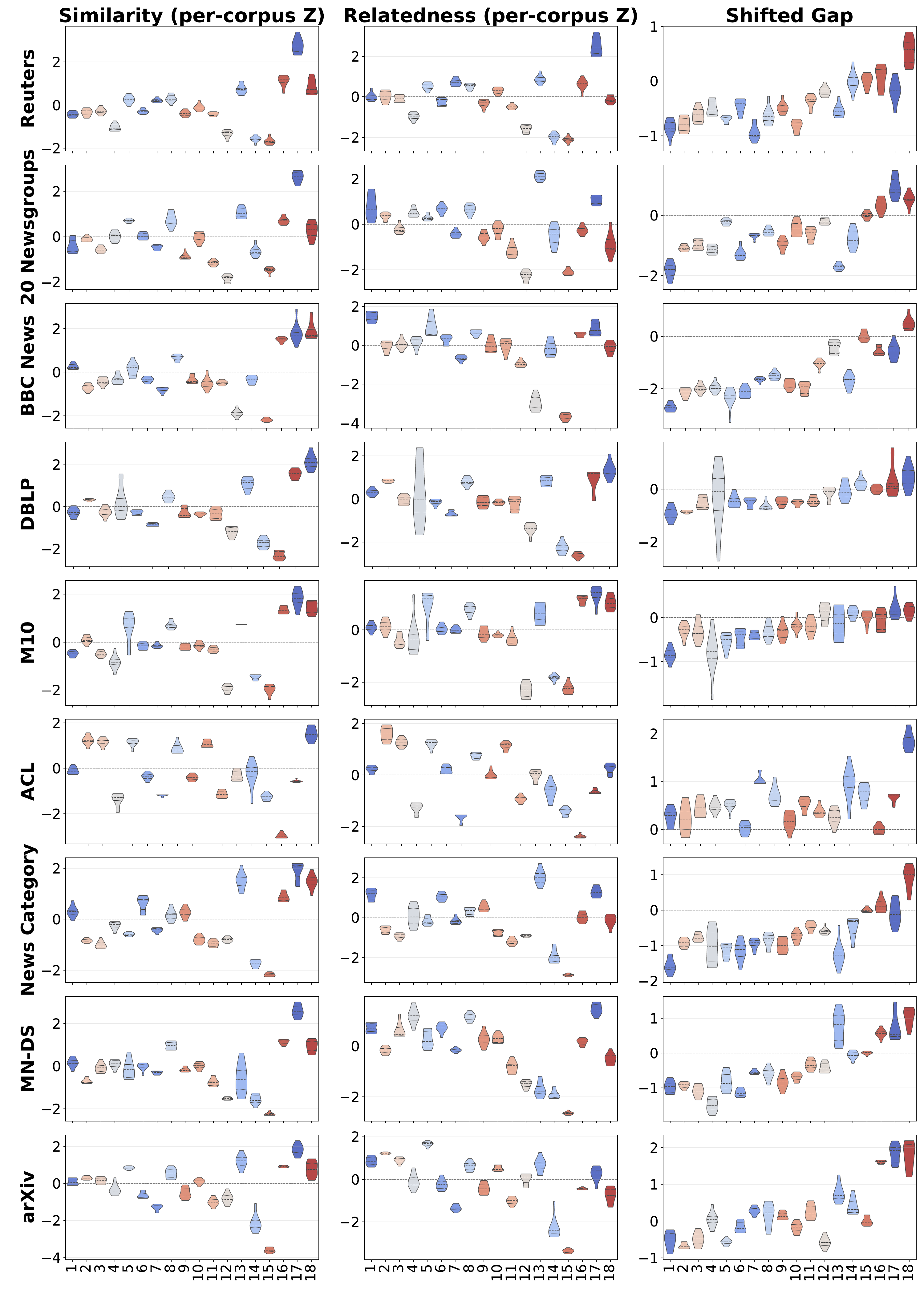}
        \caption{Per-corpus topic model atlas. Each row is one corpus; columns give the per-model similarity, relatedness, and shifted-gap distributions (per-corpus standardised, $\approx$10 runs per (model, corpus) cell; excluded cells are listed in Table~\ref{tab:omitted_models}). Models share a fixed spectrum order across all panels, so the broadly stable left-to-right color pattern reflects the cross-corpus rank concordance reported in Table~\ref{tab:kendall_w}. Models are identified on the horizontal axis by numeric id, assigned in that shared spectrum order (relatedness-leaning to similarity-leaning): \atlasAppendixLegend. See Appendix~\ref{app:model_inventory} for the full model inventory.}
        \label{fig:atlas_appendix}
    \end{figure*}

    Figure~\ref{fig:atlas_appendix} disaggregates the all-corpus spectrum of Figure~\ref{fig:AtlasReuters} into one row per corpus. Models share a fixed left-to-right order across all rows, so a stable color pattern down each column is a direct visual reading of the cross-corpus rank concordance quantified in Table~\ref{tab:kendall_w}.

\section{Reproducibility}

    \subsection{Topic Model Inventory}
    \label{app:model_inventory}
    
\begin{table*}[t]
\centering
\caption{Topic models included in the atlas. The 14 models above the mid-rule appear on all nine corpora; the four below appear on a subset only.}
\label{tab:atlas-models}
\small
\begin{tabular}{@{}p{3.0cm} p{6.8cm} l@{}}
\toprule
\textbf{Model} & \textbf{Approach} & \textbf{Reference} \\
\midrule
LDA            & Hierarchical Bayesian model; documents as mixtures of topics over vocabulary words & \citet{blei2003lda} \\[3pt]
LSI            & Truncated SVD on the term--document matrix to extract latent semantic factors & \citet{deerwester1990lsi} \\[3pt]
HDP            & Nonparametric Bayesian extension of LDA; infers the number of topics automatically & \citet{teh2006hdp} \\[3pt]
NMF            & Factorises the term--document matrix under non-negativity constraints into additive topic components & \citet{lee1999nmf} \\[3pt]
NeuralLDA      & VAE with logistic-normal prior for amortised LDA inference (AVITM, LDA variant) & \citet{srivastava2017avitm} \\[3pt]
ProdLDA        & Product-of-experts formulation in a VAE framework (AVITM, ProdLDA variant) & \citet{srivastava2017avitm} \\[3pt]
ETM (\texttt{etm\_bert})   & Topic--word distributions via inner products in embedding space; initialized with pretrained BERT word embeddings & \citet{dieng2020etm} \\[3pt]
ETM (\texttt{etm\_fresh}) & As ETM, but the embedding table is initialized randomly and trained from scratch & \citet{dieng2020etm} \\[3pt]
CombinedTM     & Combines SBERT document embeddings with a bag-of-words VAE & \citet{bianchi2021ctm} \\[3pt]
BERTopic       & Sentence-transformer embeddings clustered with HDBSCAN/UMAP; topic words via class-based TF-IDF & \citet{grootendorst2022bertopic} \\[3pt]
Contextual Top2Vec (CT2V) & Clusters contextual token embeddings to produce topics labelled in word pairs$^\dagger$ & \citet{angelov2024top2vec} \\[3pt]
ECRTM          & Embedding clustering regularisation loss enforces topic--word cluster alignment & \citet{wu2023ecrtm} \\[3pt]
Thompson \& Mimno (T\&M) & Clusters BERT contextual token representations into topic-like groups without generative modelling & \citet{thompson2020clusters} \\[3pt]
Random         & Assigns words to topics uniformly at random; lower-bound reference & --- \\

CWTM           & Encodes BERT token embeddings into word-topic vectors via weighted pooling with a mutual information objective & \citet{fang2024cwtm} \\[3pt]
TNTM           & Combines transformer-based topic embeddings with probabilistic LDA-style modelling & \citet{reuter2024tntm} \\[3pt]
FASTopic       & Dual semantic-relation reconstruction with optimal transport over document, topic, and word embeddings & \citet{wu2024fastopic} \\[3pt]
BERT-KT & Cluster mean-pooled BERT embeddings in the corpus and rerank by term frequency variants & \citet{sia2020tired} \\
\bottomrule
\end{tabular}

\vspace{2pt}
{\footnotesize $^\dagger$ Contextual Top2Vec uses word pairs as topic descriptors rather than single words. We separate the word pairs for the analysis in this paper.}
\end{table*}

    \subsection{Atlas Construction: Omitted Model--Corpus Pairs}
    \label{app:omitted}
    
    Table~\ref{tab:omitted_models} documents the (model, corpus) pairs excluded from the atlas. An audit of an earlier, longer exclusion list found most of its entries stale: those pairs produce usable topics under the current pipeline and were restored, leaving only four genuine-collapse pairs. The two failure modes are diversity collapse (all topics becoming identical or near-identical) and word-count collapse (topics containing fewer than four words). For model--corpus pairs that produced usable topics, duplicates were removed as described in the main text.

\begin{table*}[t]
\centering
\caption{Model--corpus pairs omitted from the atlas. An audit of the earlier,
longer exclusion list found most of its entries stale---those model--corpus
pairs do produce usable topics under the current pipeline and have been
restored. Only the four pairs below exhibit genuine, reproducible collapse.}
\label{tab:omitted_models}
\small
\begin{tabular}{lll}
\toprule
Model & Corpus & Failure mode \\
\midrule
CWTM          & BBC News & Diversity collapse: topics near-identical across runs \\
ETM (\texttt{etm\_fresh})& DBLP     & Diversity collapse: topic diversity $\leq 0.1$ on all runs \\
TNTM          & DBLP     & Word-count collapse: runs yield only 1--5 topics with $\leq$3 words \\
TNTM          & ACL      & Word-count collapse: runs yield only 1--4 topics with $\leq$3 words \\
\bottomrule
\end{tabular}
\end{table*}

    \subsection{Dataset Curation Pipeline}
    \label{app:datacurate}
    \subsubsection{Candidate Generation and Filtering}
    Candidate word pairs were drawn from four source families (Table~\ref{tab:pair_sources}) built over a 60,521-lemma vocabulary from the intersection of WordNet~\citep{miller1995wordnet} and BERT~\citep{devlin2019bert}. After generation, seed-overlapping and invalid pairs were removed and order-insensitive deduplication was applied, yielding the final 71,346-pair dataset.

    \begin{table*}[t]
    \centering
    \caption{Composition of the 71,346-pair training dataset by source family.}
    \label{tab:pair_sources}
    \small
    \renewcommand{\arraystretch}{1.15}
    \setlength{\tabcolsep}{4pt}
    \begin{tabular}{@{}l p{0.52\textwidth} r@{}}
    \toprule
    \textbf{Source Family} & \textbf{Sub-sources} & \textbf{Pairs} \\
    \midrule
    WordNet~\citep{miller1995wordnet} & synonyms (9.9K), cohyponyms (18.6K), hypernyms (4.6K), antonyms (1.3K) & 34,346 \\
    ConceptNet 5.7~\citep{speer2017conceptnet} & RelatedTo (5.3K), AtLocation (2.1K), PartOf (1.7K), UsedFor (0.3K), Causes (0.2K) & 9,616 \\
    Embedding neighbors & BERT top-$k$ cosine nearest neighbours (max 3/cluster, 20/word) & 14,961 \\
    Controlled negatives & far-distance (5.0K), cross-domain (3.7K), cross-POS (3.7K) & 12,423 \\
    \midrule
    \textbf{Total} & & \textbf{71,346} \\
    \bottomrule
    \end{tabular}
    \end{table*}

    \subsubsection{LLM Annotation}
    Each candidate pair was annotated independently by three large language models---DeepSeek-V4~\citep{deepseekai2026v4}, GPT-4.1-mini~\citep{openai2025gpt41}, and Qwen-Turbo~\citep{qwen2025qwenturbo}---via the prompt in Section~\ref{app:prompt}, in a single scoring pass per model ($n{=}1$). Each model first emits a two-sentence \texttt{REASON} line and then the two numeric scores (\emph{reason-first} ordering). The three per-model similarity and relatedness scores are averaged into the consensus labels used for training; their cross-model variance is retained as a per-pair uncertainty signal.

    \paragraph{API snapshot.} The exact API model identifiers used for the 71{,}346-pair production labelling are \texttt{deepseek-v4-flash} (DeepSeek-V4), \texttt{gpt-4.1-mini} (GPT-4.1-mini), and \texttt{qwen-turbo-latest} which resolves to the \texttt{qwen-turbo-2025-04-28} snapshot (Qwen-Turbo). The validation-only fourth annotator is \texttt{claude-haiku-4-5} (Claude-Haiku-4.5~\citep{anthropic2025claude45}), and the annotator-selection ablation in Section~\ref{app:annotator_ablation} additionally scores \texttt{grok-4.20-0309-non-reasoning} (Grok~\citep{xai2026grok420}), which was not used for labelling.

    \subsubsection{LLM Annotator Selection}
    \label{app:annotator_ablation}
    The annotation configuration---reasoning format, annotator model, and single vs.\ ensemble---was selected by validating against TxThmNorms~\citep{landrigan2016taxonomic}, an external benchmark of 659 word pairs carrying human-annotated taxonomic-similarity and thematic-relatedness ratings. For each configuration we score all 659 pairs and report the Pearson correlation ($r$) between LLM and human scores on each dimension. Table~\ref{tab:annotator_ablation} reports three comparisons.

    \begin{table*}[t]
    \centering
    \caption{LLM annotator selection on TxThmNorms (659 human-rated pairs). $r$ is the Pearson correlation between LLM and human scores. All single-annotator and ensemble rows use reason-first, two-sentence reasoning. Best mean $r$ per block in bold.}
    \label{tab:annotator_ablation}
    \small
    \setlength{\tabcolsep}{5pt}
    \begin{tabular}{@{}lccc@{}}
    \toprule
    Configuration & $r_{\text{sim}}$ & $r_{\text{rel}}$ & mean $r$ \\
    \midrule
    \multicolumn{4}{@{}l}{\emph{Prompt design (DeepSeek-V4)}} \\
    \quad Answer-first (scores, then reason) & 0.828 & 0.809 & 0.818 \\
    \quad Reason-first, 1 sentence & 0.873 & 0.835 & 0.854 \\
    \quad Reason-first, 2 sentences & 0.879 & 0.837 & \textbf{0.858} \\
    \quad Reason-first, 2 sent, simple prompt & 0.869 & 0.809 & 0.839 \\
    \midrule
    \multicolumn{4}{@{}l}{\emph{Single annotator}} \\
    \quad DeepSeek-V4 & 0.879 & 0.837 & \textbf{0.858} \\
    \quad Claude-Haiku-4.5 & 0.885 & 0.815 & 0.850 \\
    \quad GPT-4.1-mini & 0.845 & 0.845 & 0.845 \\
    \quad Qwen-Turbo & 0.814 & 0.801 & 0.808 \\
    \quad Grok & 0.706 & 0.848 & 0.777 \\
    \midrule
    \multicolumn{4}{@{}l}{\emph{Ensemble (mean of members)}} \\
    \quad 3-model (DeepSeek-V4 + GPT-4.1-mini + Qwen) & 0.892 & 0.880 & 0.886 \\
    \quad 4-model (+ Claude-Haiku-4.5) & 0.903 & 0.883 & \textbf{0.893} \\
    \bottomrule
    \end{tabular}
    \end{table*}

    \paragraph{Prompt design.} Eliciting a free-text rationale \emph{before} the numeric scores (\emph{reason-first}) substantially outperforms emitting the scores first (\emph{answer-first}): mean $r$ rises from 0.818 to 0.854, and the cross-dimension leakage between the two axes roughly halves. Allowing two sentences of reasoning rather than one gives a further small, consistent gain. Stripping the prompt's numeric scoring scales and worked examples down to bare definitions (\emph{simple prompt}) lowers mean $r$ to 0.839, confirming that the calibration anchors carry real signal. All subsequent runs use the full reason-first prompt with two sentences of reasoning.

    \paragraph{Annotator model.} Among single annotators (reason-first, two sentences), DeepSeek-V4 attains the highest mean correlation (0.858) and is the only near-zero-bias model; Grok~\citep{xai2026grok420} is weakest, with its similarity correlation collapsing to 0.71.

    \paragraph{Single vs.\ ensemble.} Averaging the scores of three providers (DeepSeek-V4, GPT-4.1-mini, Qwen-Turbo) beats every single annotator on \emph{both} dimensions, consistent with variance reduction. Adding a fourth provider (Claude-Haiku-4.5~\citep{anthropic2025claude45}) raises mean $r$ by only 0.007. We therefore adopt the three-model ensemble as a balance between annotation quality and cost: Claude is the most expensive provider by a wide margin, and its marginal contribution is negligible.

    \subsubsection{LLM Scoring Prompt}\label{app:prompt}
    
    Each word pair is scored by prompting DeepSeek-V4 with the system
    and user messages shown in
    Figures~\ref{fig:scoreprompt-a}--\ref{fig:scoreprompt-c}. The
    placeholders \texttt{\{word1\}} and \texttt{\{word2\}} are filled
    per pair. The model first emits a two-sentence \texttt{REASON}
    line, then the \texttt{SIM:} and \texttt{REL:} scores; only the
    two numeric scores are retained.

    \begin{promptbox}{Word-Pair Scoring Prompt (1/3)}
    \textbf{System Prompt}\\[2pt]
    {\small\itshape You are an expert linguist specializing in
    taxonomical similarity and thematic relatedness. You provide
    precise numerical scores with brief, clear reasoning.}

    \vspace{5pt}\noindent\textcolor{promptbarbg}{\rule{\linewidth}{0.4pt}}

    \needspace{4\baselineskip}%
    \vspace{3pt}\noindent\textbf{User Prompt}\nopagebreak\vspace{2pt}

    \begin{lstlisting}[style=prompt]
# Understanding Taxonomical Similarity vs. Thematic Relatedness

**Taxonomical Similarity** (taxonomic/categorical overlap):
- Measures if two words refer to the SAME TYPE of thing or share defining properties.
- For nouns: shared category membership, overlapping features.
- For adjectives: ascribe the SAME property dimension (e.g., temperature, size, mood).
- For verbs: denote the SAME kind of action, such that one could substitute for the other in many sentences without changing what physically happens.
- Synonyms have very high SIM. Words that merely co-occur but are conceptually different have low SIM.

**Thematic Relatedness** (functional / associative co-occurrence in DIFFERENT roles):
- Measures if words typically appear together in real-world scenarios playing DIFFERENT roles.
- Two words are related if they stably co-occur in the same time, place, or activity, even if they share no defining features.
- Key: functional association, complementary relationship, NOT substitution.

Important: thematic relatedness is about a typical, stable association (common co-occurrence or a clear generic relation like UsedFor / PartOf / LocatedAt / AgentAtPlace / Causes).
Do NOT inflate REL just because you can invent a scenario where both words could appear (e.g., "in a travel story"). If the association requires a constructed scenario, the pair is NOT thematically related.

## Examples by Part of Speech

### Nouns

  Synonyms / morphological variants (high SIM, low REL):
  * car / automobile         SIM 0.98   REL 0.10  (synonyms substitute, don't co-occur)
  * stair / staircase        SIM 0.80   REL 0.15  (meronymy: a stair is one step of a staircase)

  Same specific category (high SIM, moderate REL):
  * cat / dog                SIM 0.70   REL 0.50  (both pets, different species)
  * cup / mug                SIM 0.90   REL 0.50  (near-interchangeable drinking vessels)

  Complementary / functional pairs (low SIM, high REL):
  * doctor / hospital        SIM 0.10   REL 0.95  (AgentAtPlace)
  * key / lock               SIM 0.05   REL 0.95  (UsedFor, complementary)
  * monkey / banana          SIM 0.05   REL 0.90  (Eats)
  * dog / leash              SIM 0.10   REL 0.90  (UsedWith)
  * employee / employer      SIM 0.30   REL 0.75  (complementary role pair)

  Unrelated (low SIM, low REL):
  * doctor / banana          SIM 0.10   REL 0.05  (no generic role)
  * pizza / bank             SIM 0.05   REL 0.05  (no stable co-occurrence)
  * airplane / government    SIM 0.05   REL 0.05  (no functional link)

### Adjectives

  Synonyms (high SIM, low REL):
  * big / large              SIM 0.95   REL 0.10
  * happy / joyful           SIM 0.90   REL 0.15

  Same dimension, intensity difference (high SIM, moderate REL):
  * hot / warm               SIM 0.85   REL 0.40

  Co-occurring traits (different dimensions but typically co-describe the same entity; low SIM, high REL):
  * pretty / popular         SIM 0.20   REL 0.85
  * wealthy / successful     SIM 0.25   REL 0.80
  * sunny / happy            SIM 0.20   REL 0.80
  * young / energetic        SIM 0.25   REL 0.75
  * old / wise               SIM 0.30   REL 0.75

  True antonyms (same dimension, opposite poles; low SIM, low-moderate REL):
  * hot / cold               SIM 0.10   REL 0.35
    \end{lstlisting}
    \end{promptbox}
    \captionof{figure}{Scoring prompt (1 of 3): the system prompt and the start
      of the user prompt --- definitions of taxonomical similarity and
      thematic relatedness, followed by noun and adjective examples.
      Continues in Figures~\ref{fig:scoreprompt-b}
      and~\ref{fig:scoreprompt-c}.}
    \label{fig:scoreprompt-a}

    \begin{promptbox}{Word-Pair Scoring Prompt (2/3)}
    \noindent\textbf{User Prompt (continued)}\nopagebreak\vspace{2pt}

    \begin{lstlisting}[style=prompt]
### Verbs

  Synonyms (high SIM, low REL):
  * speak / talk             SIM 0.95   REL 0.15
  * begin / start            SIM 0.95   REL 0.10
  * run / sprint             SIM 0.90   REL 0.10

  Subcategory / hyponymy (high SIM, moderate REL):
  * cook / bake              SIM 0.80   REL 0.45

  Same domain, different actions (moderate SIM, moderate REL):
  * walk / run               SIM 0.65   REL 0.50

  Complementary / sequential roles (low SIM, high REL):
  * cook / eat               SIM 0.35   REL 0.85
  * write / read             SIM 0.30   REL 0.80
  * teach / learn            SIM 0.20   REL 0.85
  * buy / sell               SIM 0.20   REL 0.80
  * plant / harvest          SIM 0.40   REL 0.75

  Action-reversal pairs (high SIM, moderate REL):
  * cork / uncork            SIM 0.70   REL 0.50
  * bolt / unbolt            SIM 0.70   REL 0.50
  * dock / undock            SIM 0.70   REL 0.50

  True antonyms (moderate SIM, low-moderate REL):
  * come / go                SIM 0.30   REL 0.35
  * rise / fall              SIM 0.30   REL 0.35

  Unrelated (low SIM, low REL):
  * run / sleep              SIM 0.10   REL 0.05

## Critical Distinctions

1. **Synonyms vs. complementary pairs** form the diagonal of the SIM x REL space:
   synonyms substitute (high SIM, low REL); complementary pairs play different roles (low SIM, high REL).

2. **Antonyms come in THREE flavors. Do not conflate them.**
   - **True antonyms** (hot/cold, happy/sad, come/go, rise/fall): contrasted within the same dimension, not functionally paired. REL 0.30-0.45.
   - **Action-reversal pairs** (cork/uncork, bolt/unbolt, dock/undock): same root action, opposite direction. Sequentially linked but NOT strong co-occurrence partners. REL 0.40-0.55. SIM is higher than for true antonyms because the action mechanism is shared.
   - **Complementary role pairs** (employee/employer, creditor/debtor, buyer/seller): different roles that define each other and strongly co-occur. REL 0.65-0.80.

3. **Unrelated common words**: well-known words from different domains with no stable functional link score low on both axes (SIM and REL both 0.00-0.10). Do not invent a story to connect them.
    \end{lstlisting}
    \end{promptbox}
    \captionof{figure}{Scoring prompt (2 of 3): verb examples and the
      critical-distinctions block of the user prompt.}
    \label{fig:scoreprompt-b}

    \begin{promptbox}{Word-Pair Scoring Prompt (3/3)}
    \noindent\textbf{User Prompt (continued)}\nopagebreak\vspace{2pt}

    \begin{lstlisting}[style=prompt]
## Taxonomical Similarity Scoring Scale (non-overlapping bands)

0.95-1.00: Perfect synonyms or near-identical variants
0.85-0.95: Same specific category, very similar features
0.70-0.85: Same broad category, clear feature overlap (incl. meronymy)
0.55-0.70: Related categories with partial feature overlap
0.35-0.55: Loose categorical connection (same domain, different kind)
0.15-0.35: Different categories, minimal feature overlap
0.00-0.15: Completely different categories OR antonyms within the same dimension

**Cross-POS guidance:**
- Nouns: how similar are the *things* they refer to?
- Adjectives: do they ascribe the same property *dimension*?
- Verbs: do they denote the same kind of *action* (substitutable in many sentences)?

## Thematic Relatedness Scoring Scale (non-overlapping bands)

0.85-1.00: Extremely strong functional/thematic link, defining co-occurrence
0.70-0.85: Strong association: sequentially or complementarily paired, co-occurring adjective traits, sequential verbs
0.55-0.70: Clear contextual or role-complementary connection
0.40-0.55: Moderate: same-category co-occurrence OR action-reversal pairs
0.25-0.40: True antonyms (same dimension, opposite poles) OR very loose category overlap
0.10-0.25: Weak/indirect: shared broad domain only, no specific generic relation
0.00-0.10: No meaningful association OR synonyms (which substitute, not co-occur)

# Task
First reason about the pair in TWO sentences, then rate BOTH taxonomical similarity and thematic relatedness between:
WORD1: "{word1}"
WORD2: "{word2}"

# Rules
- Treat simple inflectional variants (plural/singular, tense) as the same lemma for similarity.
- If you cannot state a clear, generic relation in TWO sentences (e.g., UsedFor / PartOf / LocatedAt / AgentAtPlace / Causes), keep REL <= 0.20.
- Write the REASON first, then the two scores; keep the reason to two sentences.
- Use exactly 2 decimal places.

# Output format (exactly)
REASON: <two sentences explaining BOTH scores>
SIM: <0.00-1.00>
REL: <0.00-1.00>
    \end{lstlisting}
    \end{promptbox}
    \captionof{figure}{Scoring prompt (3 of 3): the taxonomical-similarity and
      thematic-relatedness scoring scales, the task block, and the
      required output format.}
    \label{fig:scoreprompt-c}

    \subsection{Neural Scorer: Architecture and Hyperparameters}
    \label{app:neural_scorer}

    \subsubsection{Architecture and Hyperparameters}

    A full summary of the scorer architecture and training hyperparameters is provided in Table~\ref{tab:scorer_config}.

    \begin{table*}[t]
    \centering
    \caption{Neural scorer architecture and training configuration. }
    \label{tab:scorer_config}
    \footnotesize
    \renewcommand{\arraystretch}{1.00}
    \setlength{\tabcolsep}{2pt}
    \begin{tabular}{@{}l l p{0.60\textwidth}@{}}
    \toprule
	    \textbf{Block} & \textbf{Setting} & \textbf{Value} \\
	    \midrule
	    Input & Word embeddings & OpenAI \texttt{text-embedding-3-large}, prompt-templated static, 3072D \\
	     & Pair features (embedding) & $[\mathbf{e}_1;\, \mathbf{e}_2;\, |\mathbf{e}_1{-}\mathbf{e}_2|;\, \mathbf{e}_1 {\odot} \mathbf{e}_2]$ ($4 \times 3072 = 12{,}288$D) \\
	     & Pair features (WordNet) & $8$D: path sim, LCH, WuP, Lin, JCN, shortest path, same category, is-synonym (binary) \\
	     & Total input dimension & $12{,}296$ \\
	    \midrule
	    Architecture & Shared trunk & $12{,}296 \rightarrow 3{,}072 \rightarrow 768$ (Linear + GELU + Dropout) \\
	     & Task heads & Similarity: $768 \rightarrow 1$ (Sigmoid); Relatedness: $768 \rightarrow 1$ (Sigmoid) \\
	     & Dropout & $p=0.4$ \\
	     & Output range & $[0,\, 1]$ for both similarity and relatedness \\
	    \midrule
	    Training & Loss & $\mathcal{L}_{\text{sim}} + \mathcal{L}_{\text{rel}}$ (masked MSE, per dimension) \\
	     & Optimizer & AdamW ($\text{lr} = 3 \times 10^{-4}$, weight decay $= 0.01$) \\
	     & LR schedule & Cosine annealing ($\eta_{\min} = 3 \times 10^{-5}$) \\
	     & Batch size & $32$ \\
	     & Early stopping & Disabled (full $100$ epochs trained) \\
	     & Train/Val/Test split & 70/15/15 \\
	     & Batch balancing & Uniform random sampling \\
	     & Training data & ${\sim}71.3$K word pairs (LLM-labeled, deduplicated) \\
	    \bottomrule
	    \end{tabular}
	    \end{table*}

    \begin{figure*}[t]
        \centering
        \includegraphics[width=\textwidth]{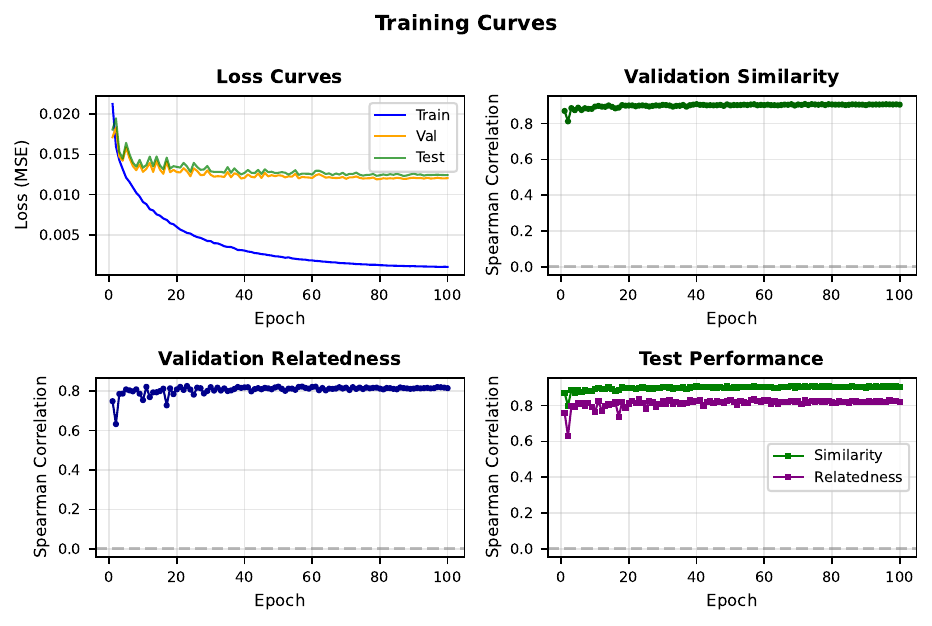}
        \caption{Training curves for the production neural scorer (\texttt{ensemble3\_rf2}, OpenAI \texttt{text-embedding-3-large} prompt embeddings + WordNet features, no gap loss). Clockwise from top-left: MSE loss (train/val/test) over epochs; validation similarity Spearman; held-out test Spearman for similarity and relatedness; validation relatedness Spearman. Validation and test Spearman correlations rise quickly and plateau by epoch~20, while train loss continues to decline---the expected train/val gap on a noisy regression label, with no degradation in held-out ranking. }
        \label{fig:training}
    \end{figure*}

    \subsubsection{POS Robustness}
    \label{app:pos_robustness}

    Because the WordNet-derived features are biased towards nouns, we evaluate whether scorer performance degrades on test pairs containing non-nouns. Table~\ref{tab:pos-robustness} reports Spearman correlation on noun--noun versus all other test pairs, where primary POS is determined by each word's most frequent part of speech among its WordNet synsets.

    \begin{table}[htbp]
    \centering
    \caption{Production scorer (\texttt{ensemble3\_rf2}) performance on noun--noun vs.\ not-noun--noun test pairs of the internal 70/15/15 split. Primary POS is determined by the most frequent part of speech among each word's WordNet synsets; pairs whose Ref or Pair word has no WordNet entry are dropped. }
    \label{tab:pos-robustness}
    \small
    \begin{tabular}{l cc}
    \toprule
    \textbf{Subset} & $\rho_{\text{sim}}$ & $\rho_{\text{rel}}$ \\
    \midrule
    Noun--noun ($n=5{,}645$) & 0.897 & 0.827 \\
    Other ($n=4{,}911$)      & 0.925 & 0.836 \\
    \midrule
    All ($n=10{,}556$)       & 0.916 & 0.836 \\
    \bottomrule
    \end{tabular}
    \end{table}

\begin{figure}[htbp]
        \centering
        \includegraphics[width=\linewidth]{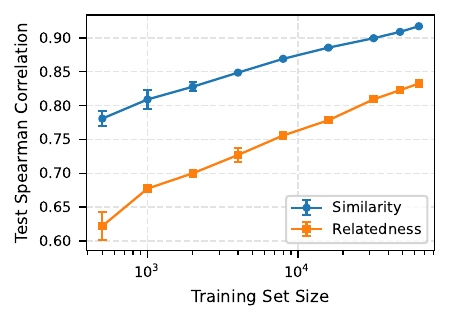}
        \caption{Scaling analysis for the production scorer (OpenAI \texttt{text-embedding-3-large} prompt embeddings + WordNet features): test-set Spearman correlation as a function of training-set size, averaged over two seeds; error bars show $\pm 1$ standard deviation. Both axes continue to improve from 500 to 64{,}000 pairs ($\rho_{\text{sim}}$: 0.78$\rightarrow$0.92; $\rho_{\text{rel}}$: 0.62$\rightarrow$0.83), with progressively smaller marginal gains beyond 16{,}000 pairs. }
        \label{fig:scale}
    \end{figure}

    \begin{samepage}
    \subsubsection{Scaling-Law Analysis}
    \label{app:scaling_law}

    Figure~\ref{fig:scale} visualizes test-set Spearman correlation as a function of training data size and highlights diminishing marginal gains at larger scales.
    \end{samepage}

    \subsection{Neural Scorer Ablation Study}
    \label{app:ablation}
    
    We evaluate the neural scorer's design choices on three complementary protocols: the synthetic internal test split, a word-level holdout of 962 pairs over 100 unseen words, and the external TxThmNorms benchmark~\citep{landrigan2016taxonomic}. We report Spearman $\rho$ for similarity (S) and relatedness (R) on each.

    \paragraph{Input embedding.} The input embedding is the scorer's most consequential design choice. We compare three families (Table~\ref{tab:embedding_ablation}): a classical, count-based word embedding (GloVe~\citep{pennington2014glove}); static embeddings derived from large language models---X2Static~\citep{gupta2021obtaining}, which is designed to distil a contextual encoder into static word vectors, alongside the Model2Vec and WordLlama distillations; and embeddings read out of sentence encoders (BGE-large and OpenAI \texttt{text-embedding-3-large}). For the sentence encoders we additionally test a \emph{prompt-augmented} variant, in which each word is wrapped in a short template before encoding. Because the embedding dimension ranges from 256 to 3072, the MLP trunk is scaled to each embedding. The prompt-augmented OpenAI embedding attains the best overall performance and is adopted for the production scorer: prompt augmentation improves on the raw single-word encoding, and a strong sentence encoder surpasses both the count-based and the distilled static embeddings. The deployed production scorer (\texttt{ensemble3\_rf2}) further retrains this configuration on the final reason-first 3-model LLM-ensemble labels (Appendix~\ref{app:annotator_ablation}); it reaches Spearman \textbf{0.905 / 0.822} on the internal test split, \textbf{0.869 / 0.798} on the word-level holdout, and \textbf{0.667 / 0.506} on external TxThmNorms---the numbers reported in Table~\ref{tab:scorer_eval} and used to score the atlas. 

    \begin{table*}[t]
    \centering
    \caption{Input-embedding ablation. Each row is one scorer trained on the 71k LLM-labeled dataset with the listed input embedding, all else fixed---a controlled comparison of embedding choices, not the deployed-scorer numbers. Spearman $\rho$ (similarity\,/\,relatedness) on the internal test split, the word-level holdout, and the external TxThmNorms benchmark; the MLP trunk is scaled to the embedding dimension. The adopted embedding and the best mean are in bold. The deployed production scorer further retrains the bold (\textbf{OpenAI te3-large, prompt}) configuration on the final reason-first 3-model ensemble labels, reaching $0.905/0.822$ (internal), $0.869/0.798$ (word-holdout), $0.667/0.506$ (TxThmNorms).}
    \label{tab:embedding_ablation}
    \footnotesize
    \setlength{\tabcolsep}{2pt}
    \begin{tabular}{@{}llccccr@{}}
    \toprule
    Embedding & Dim & MLP trunk & Internal (S/R) & Holdout (S/R) & External (S/R) & Mean \\
    \midrule
    \multicolumn{7}{@{}l}{\emph{Count-based word embedding}} \\
    GloVe 840B~\citep{pennington2014glove} & 300 & 256-64 & 0.885 / 0.811 & 0.841 / 0.811 & 0.520 / 0.437 & 0.718 \\
    \midrule
    \multicolumn{7}{@{}l}{\emph{Large-model static embeddings}} \\
    X2Static (RoBERTa-large)~\citep{gupta2021obtaining} & 1024 & 1024-512-128 & 0.877 / 0.808 & 0.841 / 0.781 & 0.532 / 0.421 & 0.710 \\
    Model2Vec (potion-base) & 512 & 256-64 & 0.751 / 0.656 & 0.754 / 0.697 & 0.466 / 0.291 & 0.603 \\
    WordLlama & 256 & 256-64 & 0.697 / 0.565 & 0.588 / 0.577 & 0.270 / 0.302 & 0.500 \\
    \midrule
    \multicolumn{7}{@{}l}{\emph{Sentence-encoder embeddings}} \\
    BGE-large, prompt & 1024 & 1024-256 & 0.850 / 0.750 & 0.830 / 0.781 & 0.516 / 0.331 & 0.676 \\
    OpenAI te3-large, raw & 3072 & 3072-768 & 0.921 / 0.833 & 0.897 / 0.822 & 0.598 / 0.343 & 0.736 \\
    \textbf{OpenAI te3-large, prompt} & 3072 & 3072-768 & 0.926 / 0.857 & 0.884 / 0.826 & 0.582 / 0.424 & \textbf{0.750} \\
    \bottomrule
    \end{tabular}
    \end{table*}

    \paragraph{Other design choices.} Table~\ref{tab:ablation_all} ablates the remaining axes. Replacing the engineered interaction features with raw concatenation $[\mathbf{e}_1;\mathbf{e}_2]$ collapses external relatedness to near zero, confirming that interaction features are essential for out-of-domain generalization. Disabling the 8-dimensional WordNet feature vector degrades every protocol by a few points, with the drop concentrated on external similarity ($.582\!\to\!.496$) while external relatedness is essentially unaffected---consistent with WordNet contributing taxonomic structure that is hardest to recover from distributional embeddings alone. The cross-attention hybrid underperforms the MLP reference, supporting the MLP design as both stronger and simpler.

    \begin{table*}[t]
    \centering
    \caption{Ablation of the remaining scorer design axes---input representation, WordNet features, and architecture---relative to the adopted configuration. Spearman $\rho$ (S,R) on the pair-heldout and word-heldout splits and the external TxThmNorms benchmark. The WordNet row is the production OpenAI-embedding scorer retrained with the 8-dimensional WordNet feature vector disabled; the input-representation and architecture rows report development-time results from the scorer design search. Each ablation isolates one axis against an otherwise-fixed configuration. }
    \label{tab:ablation_all}
    \footnotesize
    \begin{tabular}{p{1.6cm} p{2.7cm} c c c}
    \toprule
    Component & Setting & Pair-heldout (S,R) & Word-heldout (S,R) & External (Tx,Thm) \\
    \midrule
    Input & $[\mathbf{e}_1;\mathbf{e}_2]$ (no interaction) & .736, .632 & .712, .684 & .606, -.006 \\
    \midrule
    WordNet & OFF & .887, .814 & .843, .783 & .496, .420 \\
    \midrule
    Architecture & Cross-attn hybrid & .648, .578 & .631, .627 & .202, -.010 \\
    \bottomrule
    \end{tabular}
    \end{table*}

    \subsection{Corpus Statistics and Preprocessing}
    \label{app:corpora}
    
    \begin{table*}[t]
    \centering
    \caption{Summary Statistics of Text Corpora}
    \label{tab:corpus_stats}
    \small
    \setlength{\tabcolsep}{4pt}
    \begin{tabular}{llrrr}
    \toprule
    \textbf{Corpus} & \textbf{Domain} & \textbf{Documents} & \textbf{Avg Tokens} & \textbf{Vocab Size} \\ 
    \midrule
    20NewsGroups~\citep{lang1995newsweeder} & Online forum posts (20 topics) & 18,248 & 69.4 & 24,381 \\
    BBC News~\citep{greene2006bbc} & News articles (5 categories) & 2,225 & 111.0 & 2,814 \\
    Reuters-21578~\citep{lewis1997reuters21578} & Newswire (financial/economic) & 7,717 & 54.4 & 5,699 \\
    DBLP~\citep{ley2002dblp} & CS paper titles & 54,595 & 5.3 & 1,480 \\
    M10~\citep{lim2015cntm_m10} & Scientific paper titles (10 areas) & 8,355 & 5.8 & 1,591 \\
    ACL~\citep{bird2008aclarc} & NLP paper full texts & 10,551 & 1,661.0 & 8,868 \\
    \midrule
    News Category~\citep{misra2022newscategory} & News headlines (42 categories) & 39,962 & 12.3 & 11,782 \\
    MN-DS~\citep{petukhova2023mnds} & News articles (17 categories) & 10,917 & 272.5 & 29,601 \\
    arXiv~\citep{clement2019arxiv} & Scientific abstracts (20 fields) & 40,000 & 80.0 & 18,839 \\
    \bottomrule
    \end{tabular}
    \end{table*}

    The nine corpora are deliberately heterogeneous: they were compiled across several decades, span scientific writing, newswire, and informal online discussion, and range from very short titles and headlines (DBLP, M10, News Category) through abstracts (arXiv) to full-length articles (ACL, MN-DS). The lower six are established topic-modeling benchmarks; the upper three are larger, more recent collections added to test whether the atlas findings hold beyond older benchmarks. For News Category and arXiv, which are very large, we use stratified subsamples ($\approx$40K documents each) to keep per-model training cost comparable across the atlas; MN-DS is used in full.

    All nine corpora were preprocessed via a unified pipeline following the BERT-KT methodology: lowercasing, removal of punctuation, digits, and stopwords, and vocabulary pruning by minimum document frequency and token length. Contextual word embeddings were generated using \texttt{BERT-base-uncased} (layer 12) by averaging subword representations across all in-context occurrences, then projected from 768 to 300 dimensions via PCA.

    \subsection{Computational Resources}
    \label{app:compute}

    All experiments ran on a Tier-2 academic HPC system. Scorer training used a single NVIDIA A100 80~GB; atlas topic-model training used a single NVIDIA GH200 95~GiB per job. Co-occurrence models (LDA, NMF, LSI, HDP) ran CPU-only; all other model families used a single GPU. LLM-annotation jobs ran CPU-only with API latency as the bottleneck. Table~\ref{tab:compute} summarizes the dominant compute components.

    \begin{table*}[t]
    \centering
    \caption{Compute budget for the main pipeline components. Atlas training totals are approximate because per-model wall time varies by an order of magnitude across the seventeen model families and nine corpora; the slowest contributors per corpus are CombinedTM, CWTM, ECRTM, Thompson\,\&\,Mimno, and Contextual Top2Vec.}
    \label{tab:compute}
    \footnotesize
    \setlength{\tabcolsep}{4pt}
    \renewcommand{\arraystretch}{1.15}
    \begin{tabular}{@{}p{4.3cm} p{4.6cm} p{6.0cm}@{}}
    \toprule
    \textbf{Component} & \textbf{Wall-clock} & \textbf{Hardware / notes} \\
    \midrule
    LLM annotation (71{,}346 pairs $\times$ 3 providers) & ${\approx}1$~h per provider, run in parallel & 1~CPU, 5~GB RAM; API-bound. Total API spend ${\approx}\$50$--$\$90$. \\
    Neural-scorer training & 17~min & 1${\times}$A100 80GB, 100 epochs, batch 32 \\
    Scorer ablations + scaling sweep & ${\approx}9$~GPU-hours total & Same hardware as above \\
    Atlas topic-model training (1{,}567 runs across 17 topic models and a random baseline $\times$ 9 corpora $\times$ 10 seeds) & ${\approx}200$--$250$~GPU-hours total & GH200 single-GPU; CPU-only for co-occurrence models. Per-model wall time varies from $<1$~min (Random, BERT-KT, LDA) to ${\approx}1$~h (CombinedTM); Contextual Top2Vec additionally requires ${\approx}110$~GB host RAM. \\
    Atlas scoring (162 cells) & 48~min & 1${\times}$A100 80GB, 12~CPUs, 64~GB RAM \\
    Downstream evaluation (regressions, retrieval tasks, traditional metrics) & ${\leq}10$~GPU-hours & 4~CPUs, 32~GB RAM per task \\
    \bottomrule
    \end{tabular}
    \end{table*}

    \paragraph{Aggregate.} A full end-to-end re-run from raw corpora to camera-ready numbers requires roughly $\mathbf{200}$--$\mathbf{250}$~GPU-hours of single-GPU time, dominated by atlas topic-model training, plus ${\approx} \$90$ of paid LLM-API spend. Per-model wall-time and RAM details are released alongside the code.

\end{document}